\documentclass{ieeetjpreprint}
\usepackage{cite}
\usepackage{amsmath,amssymb,amsfonts}
\usepackage{algorithmic}
\usepackage{graphicx,color}
\usepackage{textcomp}
\usepackage{xcolor}
\usepackage{hyperref}
\hypersetup{hidelinks=true}
\usepackage{algorithm,algorithmic}
\def\BibTeX{{\rm B\kern-.05em{\sc i\kern-.025em b}\kern-.08em
    T\kern-.1667em\lower.7ex\hbox{E}\kern-.125emX}}
\AtBeginDocument{\definecolor{tmlcncolor}{cmyk}{0.93,0.59,0.15,0.02}\definecolor{NavyBlue}{RGB}{0,86,125}}

\def\OJlogo{}
\def\seclogo{}

\def\authorrefmark#1{\ensuremath{^{\textbf{#1}}}}

\usepackage{hyperref}
\usepackage{duckuments}
\usepackage{caption}
\usepackage{amssymb}
\usepackage{graphbox}
\usepackage{xspace}

\DeclareMathOperator*{\argmax}{\arg\!\max}

\newcommand{\Esw}{eSWEETS\textsuperscript{3}}

\newcommand{\syslongul}{\underline{G}lider \underline{A}utonomy \underline{L}ong-term Planning \underline{E}ngine\xspace}
\newcommand{\syslong}{Glider Autonomy Long-term Planning Engine\xspace}
\newcommand{\sysshort}{GALE\xspace}

\begin{document}
\currentdate{15 April, 2026}

\markboth{Online Navigation Planning for Long-term Autonomous Operation of Underwater Gliders}{Darvariu {et al.}}

\title{Online Navigation Planning for Long-term Autonomous Operation of Underwater Gliders}

\author{Victor-Alexandru Darvariu\authorrefmark{1},
        Charlotte Z. Reed\authorrefmark{1},
        Jan Stratmann\authorrefmark{2},
        Bruno Lacerda\authorrefmark{3},
        Benjamin Allsup\authorrefmark{4},
        Stephen Woodward\authorrefmark{2},
        Elizabeth Siddle\authorrefmark{4},
        Trishna Saeharaseelan\authorrefmark{4},
        Owain Jones\authorrefmark{4},
        Dan Jones\authorrefmark{4},
        Tobias Ferreira\authorrefmark{4},
        Chloe Baker\authorrefmark{4},
        Kevin Chaplin\authorrefmark{4},
        James Kirk\authorrefmark{4},
        Ashley Iceton-Morris\authorrefmark{4},
        Ryan D. Patmore\authorrefmark{4},
        Jeff Polton\authorrefmark{4},
        Charlotte Williams\authorrefmark{4},
        Christopher D. J. Auckland\authorrefmark{5},
        Rob A. Hall\authorrefmark{6},
        Alexandra Kokkinaki\authorrefmark{4},
        Alvaro Lorenzo Lopez\authorrefmark{4},
        Justin J. H. Buck\authorrefmark{4}, and
        Nick Hawes\authorrefmark{1}
}
\affil{Oxford Robotics Institute, Department of Engineering Science, University of Oxford, Oxford, UK}
\affil{Optiver, Amsterdam, The Netherlands}
\affil{Stateful Robotics, Oxford, UK}
\affil{National Oceanography Centre, Southampton, UK}
\affil{School of Environmental Sciences, University of East Anglia, Norwich, UK}
\affil{Scottish Association for Marine Science, Oban, UK}
\corresp{Corresponding author: Victor-Alexandru Darvariu (email: \texttt{victord@robots.ox.ac.uk}).}
\authornote{The work of Jan Stratmann and Bruno Lacerda was performed while affiliated with the Oxford Robotics Institute. Rob A. Hall was with the School of Environmental Sciences, University of East Anglia for this work.}

\begin{abstract}
Underwater glider robots have become indispensable for ocean sampling, yet fully autonomous long-term operation remains rare in practice. Although stakeholders are calling for tools to manage increasingly large fleets of gliders, existing methods have seen limited adoption due to their inability to account for environmental uncertainty and operational constraints. In this work, we demonstrate that uncertainty-aware online navigation planning can be deployed in real-world glider missions at scale. We formulate the problem as a stochastic shortest-path Markov Decision Process and propose a sample-based online planner based on Monte Carlo Tree Search. Samples are generated by a physics-informed simulator calibrated on real-world glider data that captures uncertain execution of controls and ocean current forecasts while remaining computationally tractable. Our methodology is integrated into an autonomous system for Slocum gliders that performs closed-loop replanning at each surfacing. The system was validated in two North Sea deployments totalling approximately 3 months and 1000 km, representing the longest fully autonomous glider campaigns in the literature to date. Results demonstrate improvements of up to 9.88\% in dive duration and 16.51\% in path length compared to standard straight-to-goal navigation, including a statistically significant path length reduction of $9.55\%$ in a field deployment. 

\end{abstract}


\maketitle

\section{INTRODUCTION}

\IEEEPARstart{O}{cean} sampling is the process of measuring indicators such as temperature, salinity, oxygen, and biomass. It is a crucial task with many applications in the sciences including biology, ecology, oceanography, climatology, and meteorology. Findings based on these observations can inform policy around societal issues such as the management of marine resources and ecosystems. Ocean sampling is inherently challenging, since it involves measuring a dynamic fluid over a range of temporal and spatial scales~\cite{rudnickUnderwaterGlidersOcean2004}.

Underwater gliders, a type of autonomous underwater vehicle (AUV), are central to addressing the ocean observation problem. In contrast to \textit{static} networks of hydrographic moorings or drifting networks of floats~\cite{roemmichAutonomousProfilingFloats2004, roemmichFutureArgoGlobal2019}, gliders allow for the sampling of \textit{dynamic} phenomena such as eddies or ocean fronts. Gliders are able to operate autonomously for months at a time and can be more energy-efficient and less carbon-intensive than ship operations~\cite{nzoc2021summary}.

Gliders form an important part of the Global Ocean Observation System, and their activity has been increasing in the last decade~\cite{testorOceanGlidersComponentIntegrated2019}. It is widely accepted that their operation and management need to become more efficient in order for organisations to control increasingly large fleets~\cite{stommelSlocumMission1989,whittFutureVisionAutonomous2020}. Standard operational procedures require input from human pilots upon the surfacing of the glider, which occurs every few hours. Pilots check environmental conditions, the progress of the previous dive, and mission objectives to decide the next control instructions. This raises challenges as piloting supervision is not always available, leading the glider to follow its most recent, possibly outdated, control instructions. Therefore, opportunities for optimising navigation are regularly lost.

\begin{figure}[t]
\begin{center}
  \includegraphics[width=\columnwidth]{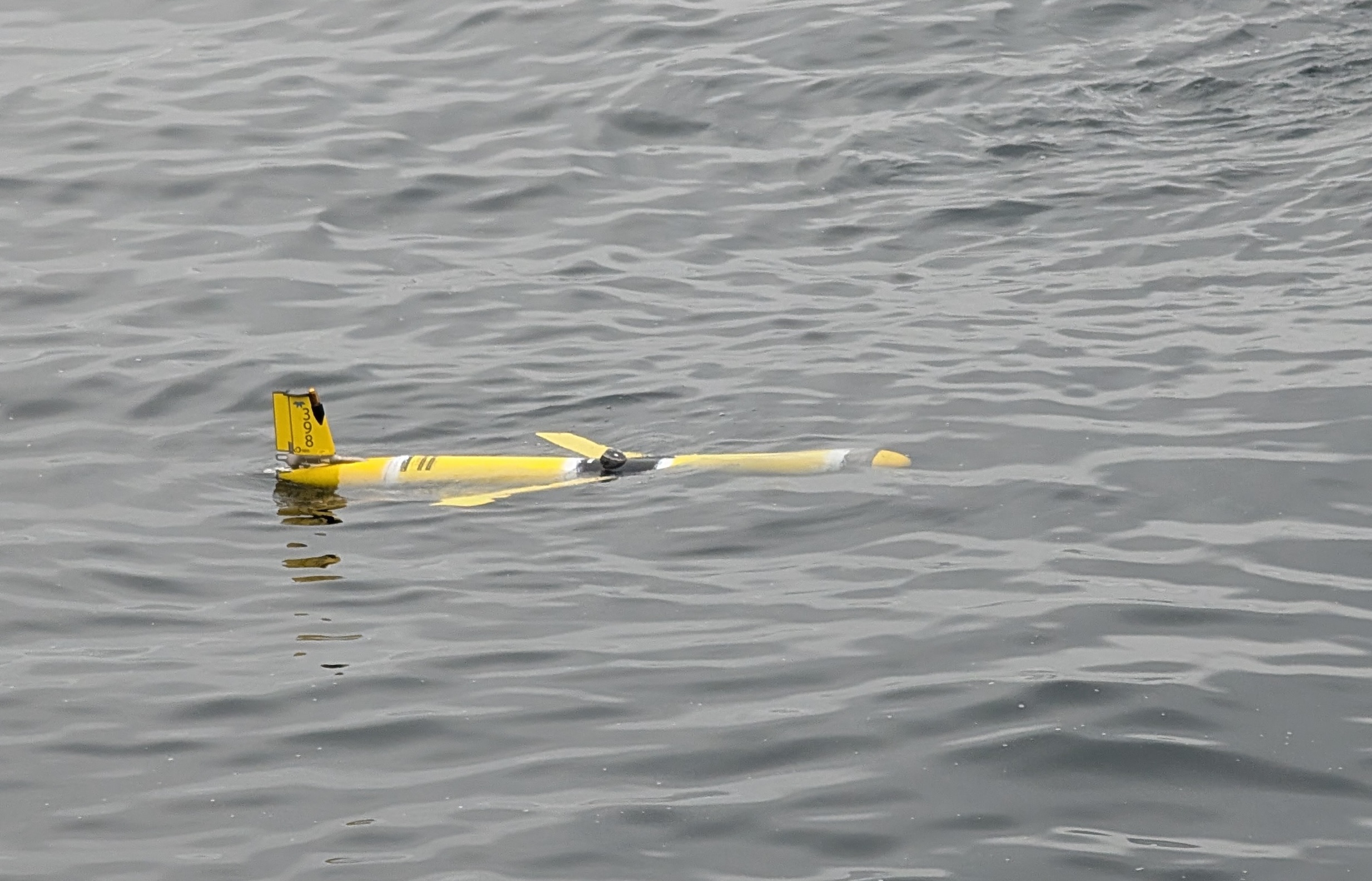}
 \caption{Slocum glider in operation during the \Esw{} field deployment. The proposed \textit{\syslong} system was used to pilot gliders autonomously in two long-term deployments totalling approximately $3$ months and $1000$ km.}
\label{fig:glider_picture}
\end{center}
\end{figure}

A promising direction to optimise navigation autonomously is to leverage navigation planning algorithms, which can plan \textit{online} at every surfacing given an up-to-date surfacing location and environmental conditions. Despite the existence of a fairly large body of work on navigation planning for underwater gliders (reviewed in Section~\ref{sec:prior_work_on_glider_mission_planning}), until now such methods have not been adopted for autonomous long-term missions. We argue that this is primarily due to the inability of such methods to account for uncertainty in ocean current forecasts and vehicle dynamics.

In this work, we formulate the glider navigation planning problem as a Markov Decision Process (MDP) in which dives correspond to actions, and in which transitions capture uncertainty in the ocean currents and glider motion. Our approach combines a computationally efficient physics-based simulator that generates possible post-dive states with an online sample-based Monte Carlo Tree Search planner. These components are integrated in the \textit{\syslongul (\sysshort)} system, which enables online, closed-loop replanning at every surfacing.

\sysshort{} was used to conduct two operational deployments in the North Sea towards refining weather forecasts and understanding the environmental impact of offshore wind farms. Our field evaluation, which exceeded $3$ months and $1000$ km in total, represents the most extensive fully autonomous deployment of online glider navigation planning to date. Results from the field evaluation and simulations demonstrate improved efficiency relative to the standard navigation approach.

\begin{figure}[t]
\begin{center}
\includegraphics[width=\columnwidth]{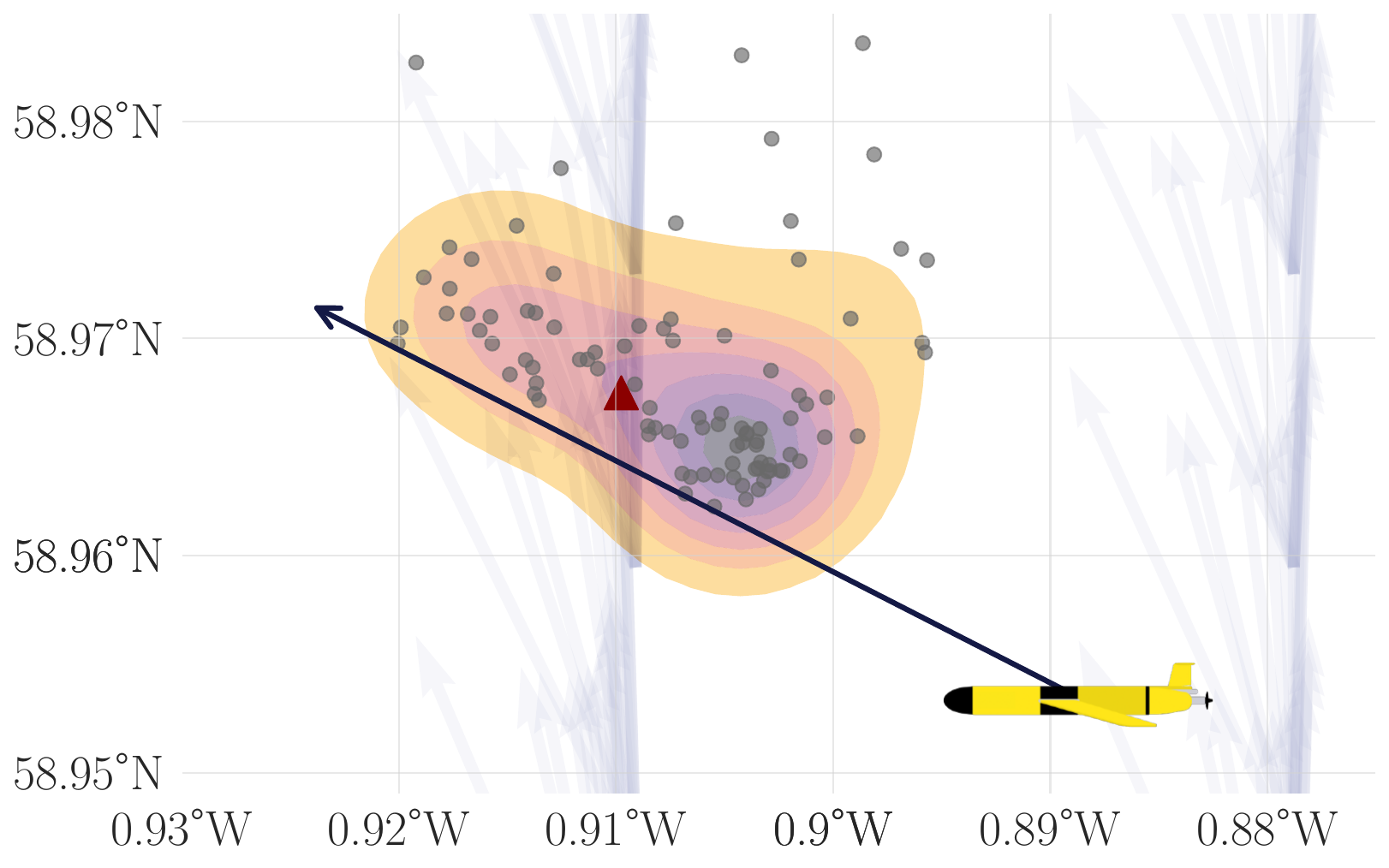} 
\end{center}
\caption{An illustration of the key inputs and outputs of the glider dive simulator. The glider icon represents the dive \textit{initial location}, the arrow indicates its \textit{direction}, and background arrows correspond to \textit{ocean current forecasts} over several depth levels. The grey circles are the outputs of the simulator indicating 
\textit{possible surfacing locations}, i.e., stochastic MDP transitions. The red triangle is the \textit{true surfacing location}.}
\label{fig:simulator}
\end{figure}

\begin{figure*}[!th]
\begin{center}
\includegraphics[width=0.85\textwidth]{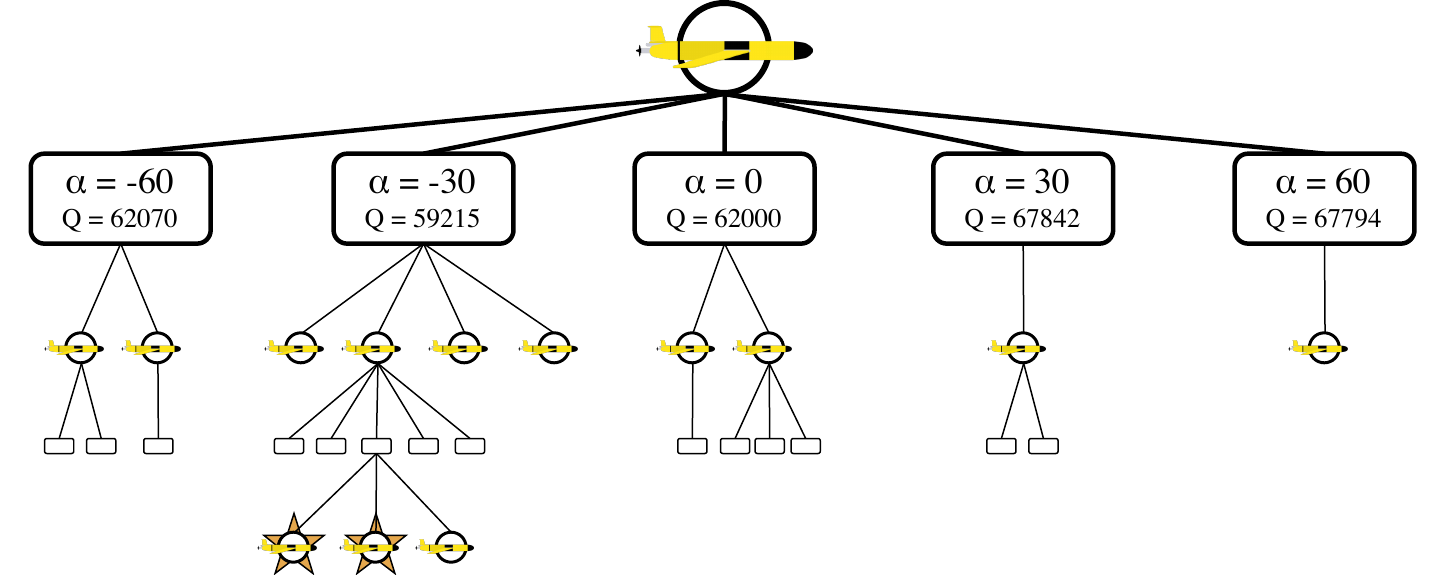}
\end{center}
\caption{An illustration of a search tree built by the planner. Circles represent state nodes and squares represent action nodes. Action nodes can have mulitple children as MDP transitions $\mathcal{T}$ are stochastic. $Q$ is the expected cost of an action. At the root, diving at $-30$ degrees relative to the goal is estimated to lead to the lowest expected cost. The search tree is several levels deep and explored asymmetrically by the planner to balance exploration and exploitation. Starred nodes denote terminal states in which the glider reaches the goal.}
\label{fig:mcts}
\end{figure*}

Our contributions are as follows:

\begin{itemize}
	\item \textbf{Markov Decision Process formulation for glider navigation planning under uncertainty.} We model multi-dive glider navigation planning as a stochastic shortest-path MDP that explicitly accounts for uncertain ocean current forecasts and glider motion.
	\item \textbf{A computationally efficient, physics-based dive simulator with data-driven calibration.} We design a simulator for MDP state transitions (Figure~\ref{fig:simulator}) that is reasonably accurate yet sufficiently fast for sample-based planning. We also propose a principled procedure for setting the simulator parameters based on a dataset of historical real-world glider dives.
	\item \textbf{An online sample-based planning approach for long-term marine autonomy.} We contribute an online sample-based planning technique based on Monte Carlo Tree Search (Figure~\ref{fig:mcts}) to solve this MDP and plan over a multi-dive horizon. The planner uses root parallelism and double progressive widening. 
	\item \textbf{The design of an autonomous online planning system for Slocum gliders.} We contribute the design of the \sysshort system, which is underpinned by the proposed methods. It is capable of controlling commercial Slocum gliders~\cite{webb2001slocum,teledyne2025slocum}.
	\item \textbf{Large-scale validation in operational conditions.} We evaluate the approach in two North Sea campaigns totalling over $3$ months and $1000$ km of autonomous operation, representing the most extensive real-world validation of online autonomous glider navigation planning to date. Compared to standard navigation, in simulation we obtain decreases of up to $9.88\%$ in dive duration and $16.51\%$ in path length on average. We also achieve a statistically significant path length reduction of $9.55\%$ in a field deployment. 
\end{itemize}

\section{UNDERWATER GLIDERS}\label{sec:gliders}

\begin{figure}[t]
\begin{center}
\includegraphics[width=\columnwidth]{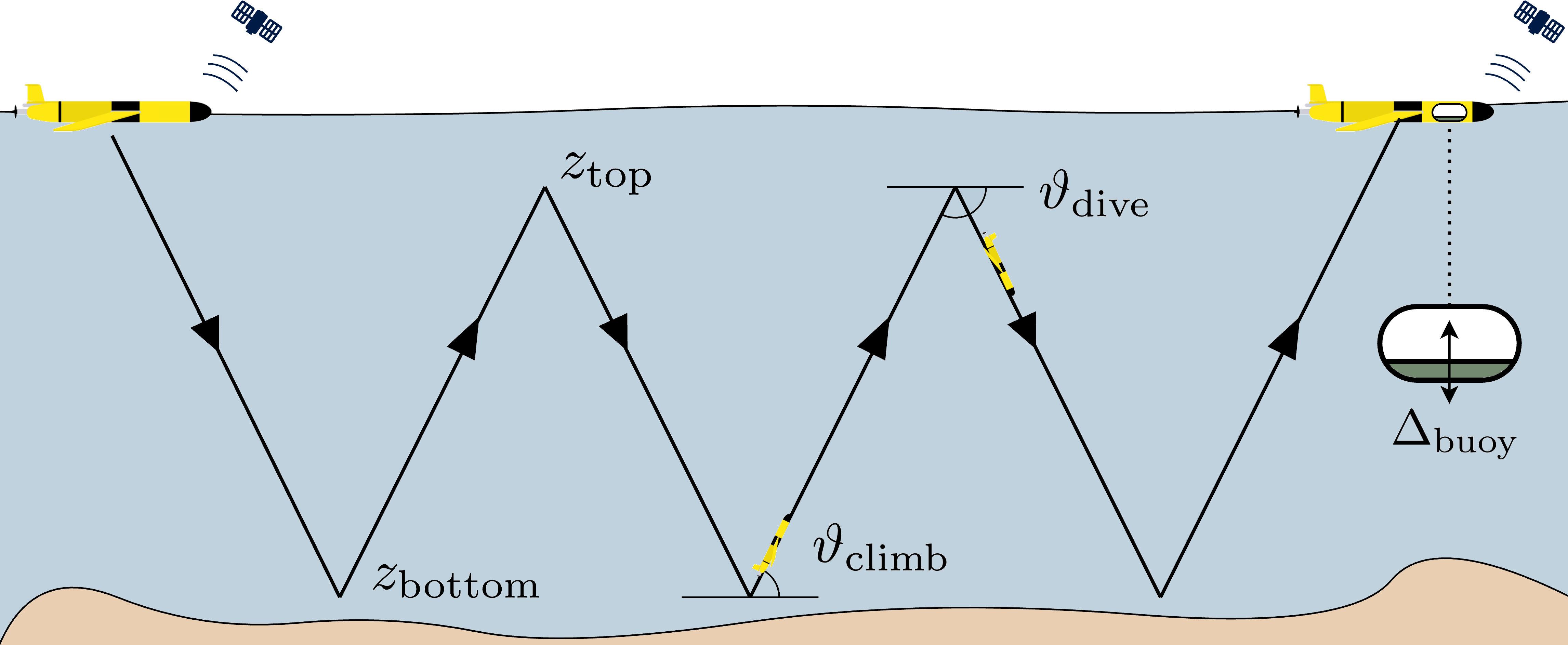} 
\end{center}
\caption{An illustration of basic glider operation and the relevant \textit{control parameters} for Slocum gliders. These include the yo floor $z_\text{bottom}$, ceiling $z_\text{top}$, climb pitch $\vartheta_\text{climb}$, dive pitch $\vartheta_\text{dive}$, buoyancy change $\Delta_\text{buoy}$, and number of yos $n_\text{yos}$ ($3$ yos are completed in this dive).  
}
\label{fig:glider_operation}
\end{figure}

An underwater glider~\cite{rudnickUnderwaterGlidersOcean2004} is a type of AUV that moves through the ocean by changing its buoyancy to glide up and down through the water column. Descent and ascent are achieved via a piston or an external bladder, changing the volume of the AUV but keeping mass constant. Movement is performed through \textit{dives} consisting of several \textit{yos}, forming a sawtooth pattern as shown in Figure~\ref{fig:glider_operation}. At the start of a yo, the volume is decreased and the vehicle descends towards the ocean floor; at the bottom of a yo, the volume is increased and the vehicle ascends. The glider's wings convert some of the vertical motion into forwards propulsion. Controlling the angle of the dive is achieved by shifting internal masses (typically the battery). Gliders are capable of operating for months at a time given their efficient passive propulsion. 

Gliders are equipped with sensors for measuring indicators such as temperature, salinity, and pressure. They also feature a GPS and satellite antenna, allowing them to transmit sampled scientific data and receive \textit{control instructions}. The instructions are comprised of two elements: the \textit{waypoints} that the glider should aim to reach, and the \textit{control parameters} that determine the behaviour of the lower-level controller on the glider during the dive. The control parameters are illustrated in Figure~\ref{fig:glider_operation}.

\begin{figure}[t]
\begin{center}
\includegraphics[width=0.9\columnwidth]{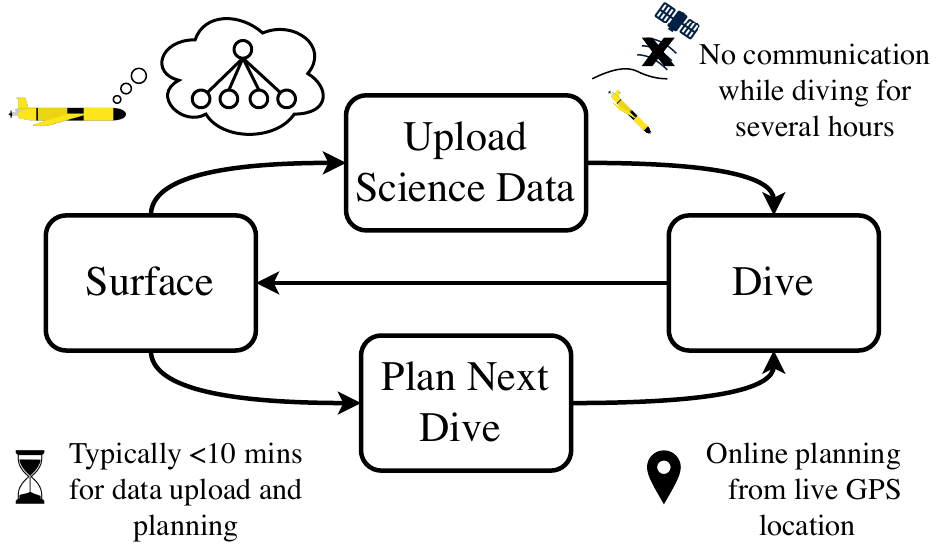} 
\end{center}
\caption{The online planning loop of the GALE system and key considerations. \textit{Upload Science Data} and \textit{Plan Next Dive} occur simultaneously while the glider is at the surface.
}
\label{fig:planning_loop}
\end{figure}

The online planning loop is illustrated in Figure~\ref{fig:planning_loop}. Control instructions are transferred over satellite while the glider is floating at the surface. The time spent floating is dictated by the amount of scientific data to be transferred and is on the order of 5-10 minutes. This implicitly defines a time budget for determining the next control instructions. An additional challenge is that transfers of control instructions are not fully reliable. Disruptions can occur during periods with adverse weather, and it is common for several transfer failures to happen in a row. The glider will use the latest set of instructions that transferred successfully, which may be significantly out of date.

Standard piloting relies on manually changing the control instructions. In between surfacings, pilots check several data sources indicating glider state and environmental factors such as wind, waves, and ocean currents in order to prepare the next set of control instructions. Given the glider surfaces several times a day for months, it is not feasible for pilots to monitor every surfacing. Despite supervision only being intermittent, estimates from recent deployments suggest that manual piloting accounts for approximately 18\% of total campaign costs, which organisations are keen on reducing.

Pilots generally set the closest waypoint such that several dives will be required to achieve it. The glider recomputes the heading towards this waypoint upon every surfacing, a mechanism we refer to as \textit{straight-to-goal (STG)}. By diving straight towards the waypoint, opportunities are regularly lost to optimise considerations such as mission duration (e.g., by setting a heading that exploits favorable currents). This motivates our investigation of online navigation planning techniques, which can adapt to changes in glider state and environmental factors after every surfacing.

\section{RELATED WORK}~\label{sec:prior_work_on_glider_mission_planning}

Glider navigation planning has been addressed under several distinct problem formulations. Works based on \textit{optimal control} focus on devising an optimal path under known flow fields. A second category of approaches relies on discrete \textit{Markov Decision Process} formulations that are solved via search or learning. Lastly, another category of works focuses on optimising coordination and coverage for \textit{fleets} of gliders, rather than planning an efficient route to a destination.

\subsection{Optimal Control}

Planning a trajectory between two waypoints was studied in~\cite{inancOptimalTrajectoryGeneration2005}, where the problem is transformed into a nonlinear programming formulation and solved via numerical optimisation. An extension for handling time-varying currents was introduced in~\cite{zhang2008optimal}. A path planning approach for estuarine environments (characterised by strong tidal currents) that can optimise for both speed and energy usage was proposed in~\cite{krugerOptimalAUVPath2007}. Ideas from fluid dynamic equations were combined with control theory in~\cite{lollaPathPlanningTime2012} to yield a general method for vehicles traveling through a variable flow field. Most recently,~\cite{aguiar2019optimizing} proposed a method for trajectory planning for AUVs with active propulsion using high-resolution ocean forecasts, applying it to an estuarine environment.

A key limitation of such path planning methods is that they assume deterministic motion and environments. As a result, they do not explicitly model uncertainty in ocean current forecasts and the movement of the glider, which can lead to compounding errors over multi-dive horizons. Furthermore, while they can achieve impressive performance in minimising cost to destination, their evaluation is typically limited to simulation. This stands in contrast to our work, which explicitly models uncertainty and evaluates performance in multi-month field campaigns.

\subsection{Search and Markov Decision Process Formulations}

Path planning using a variant of the A* algorithm was proposed in~\cite{fernandezPath2010}. To make the execution of A* tractable, the method assumes deterministic motion under a single realisation of a 2D current field and imposes a constant dive duration. The work in~\cite{al2012extending} adopts a Markov Decision Process problem formulation, considers currents across multiple depth levels, and conducts a field trial. However, this approach uses a highly limiting discretisation of the state space, which allows for offline planning using a value iteration algorithm. Uncertainty is incorporated through a Gaussian model over the summed vehicle and current velocities, yielding transition probabilities for each of the 8 surrounding cells. The approach assumes a fixed underlying current field and models uncertainty in a simplified manner, without accounting for forecast uncertainty or state-dependent execution noise. Furthermore, the field experiment is limited to two round-trip traversals between a pair of waypoints separated by $3.5$ km.

Very recent work~\cite{zhouReinforcementLearningbasedPath2026} has applied deep reinforcement learning (RL) to glider path planning. The proposed approach uses ocean current forecasts, which are paired with a simplified glider dynamics model to yield a simulation environment. The path planning policy is trained using a Soft Actor-Critic architecture for optimising a complex multi-objective reward. While improvements are demonstrated in simulation over other deep RL variants, important limitations remain: the planning occurs entirely offline and is not adjusted based on the live surfacing location of the vehicle; the method does not incorporate forecast or execution uncertainty; and the field trial is limited to a single trajectory.

\subsection{Fleet Coordination}

A control law allowing a formation of three gliders to maintain an equilateral triangle structure was formulated in~\cite{lekienGliderCoordinatedControl2008}. It can be used to determine 3-dimensional gradients of ocean properties such as temperature and salinity. A general approach for controlling a mobile sensor network such as a glider fleet was developed in~\cite{leonardCollectiveMotionSensor2007}. The latter method was successfully applied in the Adaptive Sampling and Prediction (ASAP) trial~\cite{leonardCoordinatedControlUnderwater2010}, which deployed $6$ Slocum gliders over $24$ days covering a total distance of $3270$ km. However, this required $14$ interventions from human pilots to adjust the motion patterns based on ocean forecasts. The work~\cite{hanAntColonyBasedCompleteCoveragePathPlanning2020} considered the problem of planning paths for a fleet of gliders such as to achieve complete coverage of the sampling area. Ant colony optimisation was used for altering preset paths in order to avoid obstacles. In simulation, the method yielded shorter paths with less consumed energy. We note that these approaches focus on the mission-level objectives, and are complementary to our work, which addresses the lower-level problem of navigation planning

\subsection{Summary}

Overall, our work is the first to couple uncertainty-aware online navigation planning with extended real-world deployments of ocean gliders. To our knowledge, the multi-month campaigns presented here are the most extensive demonstration to date of fully autonomous glider navigation without regular human-in-the-loop interventions.

\section{METHODS}\label{sec:methods}

\subsection{Problem Statement}

As mentioned, gliders execute dives and surface every few hours. They float while transmitting scientific information and receiving control instructions, after which they dive. During dives, they lack the ability to communicate or localise precisely, rendering the states between dives practically unobservable. Therefore, our model of the problem considers surfacings as discrete events. This enables us to approach the high-level navigation planning problem over a relatively long time horizon, while the control in between dives is handled by a lower-level controller on board the glider.

The glider's position is represented as $\mathbf{p} = (\lambda, \phi)$, where $\lambda$ and $\phi$ denote the longitude and latitude respectively. A glider state is denoted as $\mathbf{s} = (\mathbf{p}, \tau)$, where $\tau$ is a continuous timestamp in seconds. We refer to states at MDP time $t$ as $\mathbf{s}_t$, and also use $\mathbf{p}_t$ and $\tau_t$ as shorthands. We are given a start state $\mathbf{s}_0$ and goal position $\mathbf{p}_\text{goal}$ and seek to find a trajectory that minimises the total duration ${\Delta \tau}_\text{total} = \tau_T - \tau_0$ for surfacing within a radius $\rho$ of the goal at the terminal timestep $T$.

Glider navigation is strongly impacted by \textit{currents}. We represent them as the vector $\mathbf{q}(\lambda, \phi, z, \tau) = [u,v,w]^\top$ denoting the velocity in the eastward, northward, and downward components respectively. Here, $z$ represents the depth. The \textit{bathymetry} $\mathbf{b}(\lambda, \phi)=z_\text{bathy}$ identifies the maximum ocean depth at each position and is relevant since gliders typically inflect when approaching the ocean floor. In contrast to currents, which we cannot measure or predict fully accurately, we may assume bathymetry is known and fixed on the timescales of the considered planning problem.

Each instance $I$ of the glider navigation planning problem is therefore defined as the tuple $I = (\mathbf{p}_0, \tau_0, \mathbf{p}_\text{goal}, \rho, \mathbf{q}(\cdot), \mathbf{b}(\cdot))$. 

\subsection{Markov Decision Process Formulation}~\label{subsec:mdp_formulation}

MDPs are a widely employed formalisation of decision-making~\cite{kolobov2012planning,sutton2018reinforcement}. We define an MDP as the tuple $(\mathcal{S,A,T,C,G})$, where $\mathcal{S}$ is the \textit{state space} defining all configurations the agent may find itself in, $\mathcal{A}$ is the \textit{action space} defining valid controls, $\mathcal{T}: \mathcal{S} \times \mathcal{A} \times \mathcal{S} \to [0,1]$ denotes the \textit{transition function} satisfying $\sum_{a \in \mathcal{A}}{\sum_{s' \in \mathcal{S}}{\mathcal{T}(s,a,s')}} = 1\ \forall s \in \mathcal{S}$, and $\mathcal{C} : \mathcal{S} \times \mathcal{A} \times \mathcal{S} \to [0, \infty)$ is a \textit{cost function}. Also let $\mathcal{G} \subseteq \mathcal{S}$ denote a set of absorbing \textit{goal states} s.t. $\forall s_g \in \mathcal{G} \land s' \notin \mathcal{G}$ we have $\mathcal{T}(s_g, a, s_g) = 1$, $\mathcal{T}(s_g, a, s') = 0$, and $\mathcal{C}(s_g, a, s_g) = 0$. This class of models is known as \textit{stochastic shortest-path (SSP)} MDPs. A deterministic, stationary, Markovian policy is a mapping $\pi : \mathcal{S} \to \mathcal{A}$. The \textit{value function} of such a policy for a state $s$ is defined as $V^\pi(s) = \mathbb{E} [\sum_{t'=0}^{\infty}{C_{t'+t}^{\pi_{s}}} ]$, where $C_{t'+t}^{\pi_{s}}$ is a random variable denoting the cost incurred as a result of executing policy $\pi$ starting in state $s$ for all $t$. We seek to find the optimal policy $\pi^*$ which satisfies $V^*(s) = \min_{\pi} V^{\pi}(s)$, i.e., the policy that achieves the minimal cost in expectation at each state.

To formulate glider navigation planning as an SSP MDP, we must define each element of $(\mathcal{S,A,T,C,G})$. We consider each dive as a discrete decision, hence the MDP time $t$ advances by $1$ with each dive.

\noindent \textbf{States.} The state space $\mathcal{S}$ is continuous and comprises all positions $\mathbf{p}$ and times $\tau$ the glider may be in. In practice, these values are bounded: $\lambda, \phi \in [-180^{\circ}, 180^{\circ}]$ while $\tau \geq \tau_0$ and is capped by the maximum planning horizon (typically a few days).

\begin{figure}[t]
\begin{center}
  \includegraphics[width=\columnwidth]{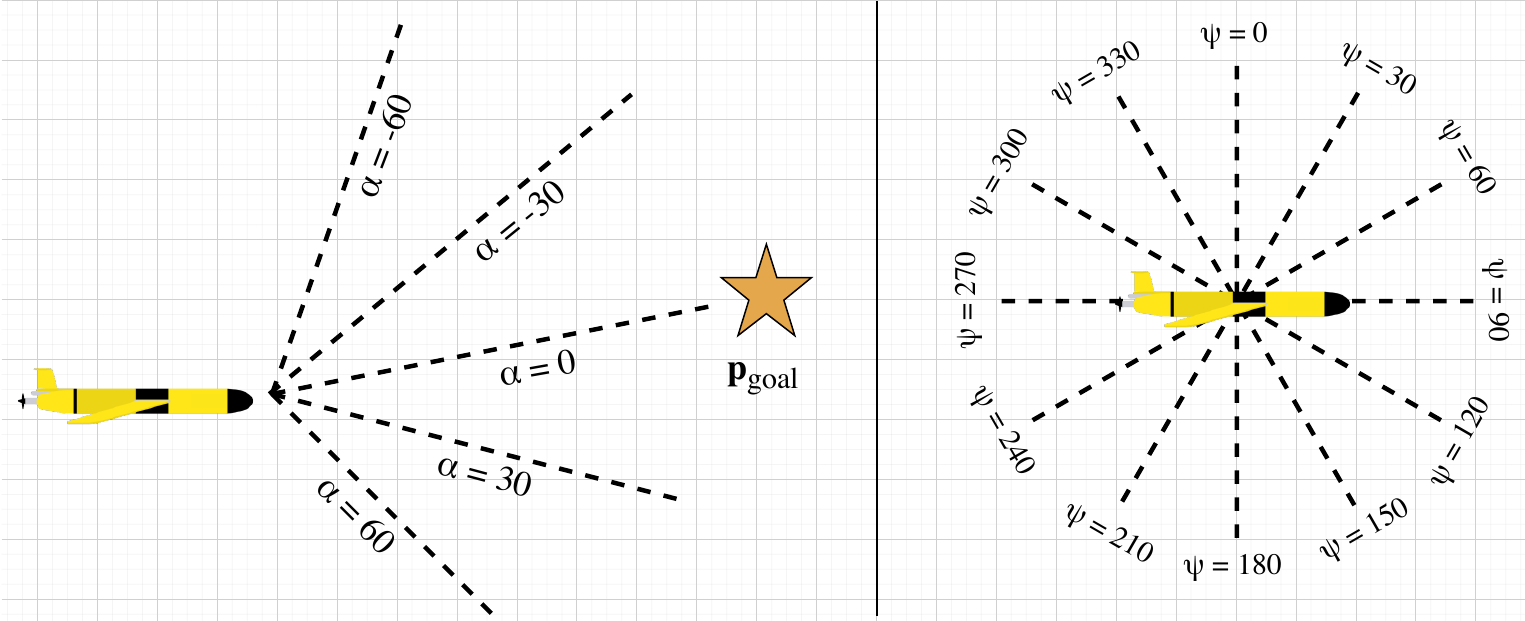}
 \caption{Illustration of the proposed action space formulation (left), which uses bearings $\alpha$ that are relative to the goal. This leads to a lower branching factor than deciding headings $\psi$ directly (right) while retaining relevant actions.}
\label{fig:action_space}
\end{center}
\end{figure}

\noindent \textbf{Actions.} Actions $\mathcal{A}$ are defined as tuples $\mathbf{a} = (\alpha, \mathbf{c})$, where $\alpha$ denotes the dive direction, while $\mathbf{c}$ is itself a tuple consisting of the dive control parameters. 
We opt to discretise the dive direction component of the action space. This allows us to incorporate the knowledge that small increments of change in heading are largely irrelevant given the noise in the system, and avoids the challenges of working directly with a continuous action space. Furthermore, we leverage the observation that glider headings $\psi$ are set relative to the bearing $\beta$ to the goal position $\mathbf{p}_\text{goal}$, as illustrated in Figure~\ref{fig:action_space}. We note the distinction between the bearing to the goal (a geometric quantity) and the vehicle heading (a control variable). Setting $\psi=\beta$ corresponds to aiming the glider \textit{straight-to-goal} which, as previously mentioned, is the default mechanism by which Slocum gliders set their heading towards a distant waypoint. Instead, we may wish to set a heading that differs from the bearing in order to account for the effect of the currents. Therefore, we allow as actions a discrete set of relative bearings $\alpha = \beta - \psi$. Assuming a north-east-down reference frame, positive values of $\alpha$ correspond to diving clockwise relative to the goal, while negative values correspond to diving anti-clockwise. This preserves a lower branching factor compared to deciding the heading directly, while prioritising actions that are more likely to lead to progress towards the goal.

\noindent \textbf{Transitions.} Transitions $\mathcal{T}$ govern the probability of the glider surfacing in state $\mathbf{s}_{t+1}$ after diving from state $\mathbf{s}_t$ using action $\mathbf{a}_t$. In this problem, transitions are impacted by uncertain currents $\mathbf{q}$ and glider motion, and cannot be modelled fully accurately. Transitions are therefore stochastic and highly uncertain. To devise navigation plans, we thus require access to a sample-based \textit{simulator} that can generate samples of likely surfacing states. This component is described in depth in Section~\ref{subsec:simulator}. 

\noindent \textbf{Costs.} The cost function $\mathcal{C}$ we consider is $\mathcal{C}(\mathbf{s}_t,\mathbf{a}_t, \mathbf{s}_{t+1}) = \tau_{t+1} - \tau_t$, i.e., we incur a cost equal to the duration of the dive. Alternative cost definitions may also account for the energy usage of each set of controls; however, in our considered experimental setup, this is approximately constant with respect to time.   

\noindent \textbf{Goals.} The goals $\mathcal{G}$ are defined as $\mathcal{G} = \{ \mathbf{s} \in \mathcal{S} \ |\ d(\mathbf{p}, \mathbf{p}_\text{goal}) \leq \rho  \}$, where $\rho$ is a tolerance radius and $d$ is geodesic distance.
 
Regarding the impact of currents on transitions $\mathcal{T}$, as we cannot accurately measure and predict currents across the entire water column, we must rely on forecasts $\mathbf{\hat{q}}$. Many ocean models exist, which vary in accuracy as well as spatial and temporal resolutions. Such models employ discretisation by splitting coordinates, depth, and time into intervals. Concretely, as we focus our deployment and evaluation in the North Sea region, we make use of the AMM15~\cite{graham2018amm15} model, which provides forecasts at hourly intervals over grid cells of $1.5 \times 1.5$ km and 33 levels whose coarseness increases with depth. We obtain these forecasts via the Met Office marine data service~\cite{metoffice2025dataservice}. These forecasts have a horizon of 5 days. This model does not make vertical velocities available in the publicly available outputs; therefore, we assume $w=0$, i.e., $\hat{\mathbf{q}}(\lambda, \phi, z, \tau) = [u,v,0]^\top$, which is a reasonable assumption in tidally dominated shelf seas.

\subsection{Custom Glider Dive Simulator}\label{subsec:simulator}

Our approach requires a \textit{simulator} that can generate samples from the transition function $\mathcal{T}$. Recall that, as described in Section~\ref{sec:gliders}, control instructions are decided within a fixed time budget of 5-10 minutes while the glider is floating at the surface. Given that the number of trials that can be run in a fixed time budget strongly influences MCTS performance~\cite{browneSurvey2012}, the simulator must balance the accuracy of sampled transitions with execution speed. Constructing a search tree over a horizon of several dives is crucial for going beyond reactive current correction.

The state-of-the-art open source simulator for Slocum gliders is \texttt{glidersim}~\cite{merckelbach2025glidersim}, whose detailed approach takes into account the actuator behaviour and hydrodynamics of Slocum gliders~\cite{merckelbach2016depth,merckelbachDynamicFlightModel2019}. However, this comes at the expense of costly runtime. The method also does not account for the unknowns surrounding the current forecasts and therefore it is unsuitable for use in our approach. Thus, we set out to devise a custom simulator that is more appropriate for sample-based planning methods.

To do so, we leverage the following observations. Firstly, the majority of large-scale, publicly available ocean models (including AMM15) do not include vertical velocities in the published model outputs. As such, we cannot use forecasts to account for changes in the glider's descent and ascent rate, and we can treat them as approximately constant. While vertical density gradients (stratification) can also influence vertical motion in practice, capturing these effects would require a more detailed hydrodynamic model, which we leave to future work. Secondly, the glider moves in a largely steady state (i.e., constant speed and attitude relative to the water) when performing the sawtooth motion. This is not the case at the bottom and top of each yo, when the onboard controller changes the glider pose and velocity. Nevertheless, these transient phases can be considered negligible in terms of the impact on the overall trajectory, as demonstrated by prior studies~\cite{wangSteadyMotionUnderwater2022,wuCombinationOptimizationMethod2025}. This motivates our simulation approach that chains together steady-state segments.  

\begin{algorithm}[!t]
\caption{Custom Glider Dive Simulator}
\label{alg:glidersim}
\begin{algorithmic}[1]

\STATE \textbf{Params:} $\Theta = \{\theta_{\text{drag\_i}}, \theta_{\text{drag\_j}}, \theta_\text{curr\_mag}, \theta_\text{curr\_dir}, \theta_\text{curr\_min},$ \\
\qquad \qquad \qquad\ \  $\theta_\text{motion\_mag}, \theta_\text{motion\_dir}, \theta_\text{motion\_min}\}$ \\
\vspace{1mm}
\STATE \textbf{Function} \texttt{SimulateDive}($\mathbf{s}, \mathbf{a}, \beta, \hat{\mathbf{q}}, \mathbf{b}$):
\STATE \quad $\mathbf{s}' \leftarrow \mathbf{s}$, \texttt{halfYosDone} $\leftarrow 0$
\STATE \quad \textbf{while} \texttt{halfYosDone} $ < 2 * \mathbf{a}.\mathbf{c}.n_\text{yos}$ \textbf{do}
\STATE \qquad \textbf{if} \texttt{halfYosDone} $\text{mod}\ 2 = 0$ \textbf{then}
\STATE \qquad\quad $\mathbf{s}' \leftarrow$ \texttt{SteadyDown}($\mathbf{s}', \mathbf{a}, \beta, \hat{\mathbf{q}}, \mathbf{b}$)
\STATE \qquad \textbf{else}
\STATE \qquad\quad $\mathbf{s}' \leftarrow$ \texttt{SteadyUp}($\mathbf{s}', \mathbf{a}, \beta, \hat{\mathbf{q}}, \mathbf{b}$)
\STATE \qquad \texttt{halfYosDone} += $1$
\STATE \quad \textbf{return} $\mathbf{s}'$
\vspace{2mm}
\STATE \textbf{Function} \texttt{SteadyDown}($\mathbf{s}, \mathbf{a}, \beta, \hat{\mathbf{q}}, \mathbf{b}$):
\STATE \quad $\psi \leftarrow$ \texttt{HeadingFromAction}($\mathbf{a}.\alpha, \beta$) 
\STATE \quad $s, w_g \leftarrow$ \texttt{GliderFlightModel}($\mathbf{a}.\mathbf{c}$)
\STATE \quad $u_g, v_g \leftarrow s \cdot \sin{\psi}, s \cdot \cos{\psi}$
\STATE \quad $u_g, v_g \leftarrow$ \texttt{ApplyMotionNoise}($u_g, v_g$)
\STATE \quad $\mathbf{g} \leftarrow [\mathbf{p}, z, \tau]^\top$
\STATE \quad $\dot{\mathbf{g}} \leftarrow [u_g, v_g, -w_g]^\top$
\STATE \quad \textbf{while} $\mathbf{g}[2] > \min{(\mathbf{a}.\mathbf{c}.z_{\text{bottom}}, \mathbf{b}(\mathbf{g}))}$ \textbf{do}
\STATE \qquad $\mathbf{g} \leftarrow$ \texttt{ToNextCell}$(\hat{\mathbf{q}}, \mathbf{g}, \dot{\mathbf{g}}, \mathbf{a}.\mathbf{c}.z_{\text{bottom}})$
\STATE \quad $(\mathbf{p}', \tau') \leftarrow (\mathbf{g}[1], \mathbf{g}[3])$
\STATE \quad \textbf{return} $(\mathbf{p}', \tau')$

\vspace{2mm}
\STATE \textbf{Function} \texttt{ToNextCell}($\hat{\mathbf{q}}, \mathbf{g}, \dot{\mathbf{g}}, z_{\text{lim}}$)
\STATE \quad $u, v, \_ \leftarrow \hat{\mathbf{q}}(\mathbf{g}$)
\STATE \quad $\tilde{\mathbf{q}}(\mathbf{g}) \leftarrow $ \texttt{GetNoisyCurrents}($u, v$) 
\STATE \quad $\dot{\mathbf{g}}_{\text{true}} \leftarrow$ \texttt{ApplyCurrAccel}($\dot{\mathbf{g}}, \tilde{\mathbf{q}}(\mathbf{g}), \theta_{\text{drag\_i}}, \theta_{\text{drag\_j}})$
\STATE \quad $\Delta \tau \leftarrow$ \texttt{TimeToNextCell}$(\hat{\mathbf{q}}, \mathbf{g}, \dot{\mathbf{g}}_{\text{true}}, z_{\text{lim}})$
\STATE \quad $\mathbf{g} \leftarrow \mathbf{g} + \dot{\mathbf{g}}_{\text{true}} \cdot (\Delta \tau + \varepsilon)$
\STATE \quad \textbf{return} $\mathbf{g}$

\vspace{2mm}
\STATE \textbf{Function} \texttt{SteadyUp}($\mathbf{s}, \mathbf{a}, \beta, \hat{\mathbf{q}}, \mathbf{b}$):
\STATE \quad /* Similar to \texttt{SteadyDown} but with $+w_g$ in $\dot{\mathbf{g}}$ and condition $\mathbf{g}[2] < \mathbf{a}.\mathbf{c}.z_{\text{top}}$ */

\vspace{2mm}
\STATE \textbf{Function} \texttt{TimeToNextCell}($\hat{\mathbf{q}}, \mathbf{g}, \dot{\mathbf{g}}_{\text{true}}, z_{\text{lim}}$)
\STATE \quad /* Calculates time until next forecast cell $\hat{\mathbf{q}}$ is reached using
geographical distances and glider speed $\dot{\mathbf{g}}_{\text{true}}$. */

\vspace{2mm}
\STATE \textbf{Function} \texttt{GetNoisyCurrents}($u, v$):
\STATE \quad $|q| \leftarrow \sqrt{u^2 + v^2}, \alpha_\text{curr} \leftarrow \text{atan}(v, u)$
\STATE \quad \texttt{magN}$\ \sim \mathcal{N}(0, \theta_\text{curr\_mag})$, \texttt{dirN}$\ \sim \mathcal{N}(0, \theta_\text{curr\_dir})$ 
\STATE \quad $|q|$ += \texttt{magN}, $\alpha_\text{curr} $ += \texttt{currN}
\STATE \quad $|q| = \max(|q|, \theta_\text{curr\_min})$
\STATE \quad \textbf{return} \texttt{polarToVec}($|q|, \alpha_\text{curr}$)

\vspace{2mm}
\STATE \textbf{Function} \texttt{ApplyMotionNoise}($u, v$):
\STATE \quad /* Similar to \texttt{GetNoisyCurrents} but with generating\texttt{magN}, \texttt{dirN} using random walks
with parameters $\theta_\text{motion\_mag}, \theta_\text{motion\_dir}$ and minimum speed $\theta_\text{motion\_min}$. */

\end{algorithmic}
\end{algorithm}

The pseudocode for our glider dive simulator is given in Algorithm~\ref{alg:glidersim}. The \textit{inputs} to the simulator are the glider state $\mathbf{s}$, action $\mathbf{a}$, bearing $\beta$, current forecasts $\hat{\mathbf{q}}$ and bathymetry $\mathbf{b}$. The \textit{output} of the simulator is the post-dive state $\mathbf{s}'$. We note that, unlike in the MDP states described in Section~\ref{subsec:mdp_formulation}, in which the glider is always at the surface, the internal glider states in the simulator also contain a depth level. We denote this internal state as $\mathbf{g}$. Querying the forecast and bathymetry are denoted $\hat{\mathbf{q}}(\mathbf{g})$ and $\mathbf{b}(\mathbf{g})$. 

The main loop of the algorithm (\texttt{SimulateDive}) simulates the characteristic seesaw-like motion by chaining together upwards and downwards steady-state phases (\texttt{SteadyUp} and \texttt{SteadyDown}) until the number of yos to perform is exhausted. The \texttt{SteadyDown} method simulates a downward steady-state phase by repeatedly propagating the glider to the next forecast cell. The \texttt{SteadyUp} method does the same except with the glider moving upwards. The general idea is that, as the forecast currents are representative for a whole grid cell, the glider state can be propagated linearly within that cell until the cell border is hit (\texttt{TimeToNextCell}). Glider horizontal speed $s \in \mathbb{R}^+$ and vertical speed (rate of change of depth) $w_g \in \mathbb{R}^+$ are calculated (\texttt{GliderFlightModel}) using the approach of Merckelbach \textit{et al.}~\cite{merckelbachDynamicFlightModel2019}. This takes into account the control parameters in $\mathbf{c}$ as well as water properties. The flight model parameters are fit to historical glider data from the region of interest. Acceleration is applied to the glider speeds as determined by the current forecast and drag coefficients $\theta_{\text{drag\_i}}, \theta_{\text{drag\_j}}$ that are part of the simulator parameters.

By default, transitions $\mathcal{T}$ generated by this simulator would be deterministic. However, as we would like the produced navigation plans to be robust to uncertainty in the forecast and glider motion, we inject \textit{current noise} and \textit{motion noise} in order to yield several plausible samples of transitions given the same inputs. 

We use a simple Gaussian noise model for the current noise (\texttt{GetNoisyCurrents}). Independent noise is added to the magnitude and direction by sampling from Gaussian distributions with mean $0$ and configurable standard deviations $\theta_\text{curr\_mag}$, $\theta_\text{curr\_dir}$. These perturbations are applied consistently for all forecasts $\hat{\mathbf{q}}$ upon the initialisation of the simulator with a particular random seed, corresponding to a systematic bias in the forecast. 

To implement motion noise (\texttt{ApplyMotionNoise}), we update the glider velocity with a noise model suggested by glider pilots. Namely, we use a random walk to reflect the noise of the internal dead-reckoning characterised by a very inaccurate state estimation and other local perturbations. We simulate a symmetric random walk over integers with absorbing barriers for which the probability of the value decreasing, staying put, or increasing is uniform. Separate walks are used for the magnitude and direction. The values of the random walk states are multiplied with the $\theta_\text{motion\_mag}$ and $\theta_\text{motion\_dir}$ noise parameters then added to the glider magnitude and direction.

The simulator parameters $\Theta$, enumerated in Line 1 of Algorithm~\ref{alg:glidersim}, control the impact of the currents on glider speed and the amount of noise that is injected into the simulator. As these parameters are challenging to establish a priori and depend significantly on the deployment vehicle, location, and conditions, we propose a procedure to tune them using past glider data.

\subsection{Simulator Parameter Tuning}\label{subsec:sim_tuning}

Choosing the simulator parameters $\Theta$ appropriately is of crucial importance for ensuring the fidelity of the generated transitions from $\mathcal{T}$ and, consequently, the effectiveness of the generated navigation plans. We denote an instance of the simulator parameterised with a particular $\Theta$ as $\mathcal{T}_\Theta$. 

To enable tuning $\Theta$, we assume availability of a dataset $\mathcal{D} = \{ \mathbf{x}^{(i)} \}_{i=1}^N$ of historical dives with entries $\mathbf{x}^{(i)}$ of the form $((\mathbf{s}^{(i)}, \mathbf{a}^{(i)}, \beta^{(i)}, \hat{\mathbf{q}}^{(i)}, \mathbf{b}^{(i)}), \mathbf{s}_\text{true}^{(i)})$, comprising the conditions prior to starting a dive and the true post-dive state after the surfacing of the glider. For simplicity, we also assume that all parameters $\mathbf{c}$ except $n_\text{yos}$ and $z_\text{bottom}$ are held constant within a dataset used to fit the simulator parameters.

We focus on optimising the generated simulator positions and dive durations relative to the ground truths $\mathbf{p}^{(i)}_\text{true}$ and $\Delta\tau^{(i)}_\text{true}$ respectively, where $\Delta\tau = \tau' - \tau$. For each $\mathbf{x}^{(i)}$, we draw $M$ samples of future states $\mathbf{s}^{(i,1)}, \mathbf{s}^{(i,2)}, \dots, \mathbf{s}^{(i,M)} \sim \mathcal{T}_\Theta(\mathbf{s}, \mathbf{a}, \beta, \hat{\mathbf{q}}, \mathbf{b})$, recording the positions $\mathcal{P}^{(i)} = \{\mathbf{p}^{(i,1)}, \mathbf{p}^{(i,2)}, \dots, \mathbf{p}^{(i,M)}\}$ and the durations $\mathcal{I}^{(i)} = \{ \Delta\tau^{(i,1)}, \Delta\tau^{(i,2)}, \dots, \Delta\tau^{(i,M)}\}$. We subsequently fit a per-dive Kernel Density Estimator (KDE) for simulator positions with a Gaussian kernel $\hat{p}^{(i)}_\Theta = \text{KDE}(\mathcal{P}^{(i)}; h)$ and bandwidth $h$, and a Gaussian distribution $\widehat{\Delta\tau}^{(i)}_\Theta \sim \mathcal{N}(\mathcal{I}^{(i)} ; \mu^{(i)}, \sigma^2)$ for dive durations with a fixed $\sigma$.

Evaluating the probabilities $\hat{p}_{\Theta}^{(i)} (\mathbf{p}^{(i)}_\text{true})$ and $\widehat{\Delta\tau}_{\Theta}^{(i)} (\Delta\tau^{(i)}_\text{true})$ quantifies how well the simulator parameters $\Theta$ fit a particular dive. Furthermore, we also penalise the spread $\sum_{j=1}^M{ (\mathbf{p}^{(i, j)} - \overline{\mathbf{p}^{(i,j)}})}^2$ of the positions relative to their mean. Without this penalty, the optimisation procedure tends to favour parameter configurations with overly diffuse transition distributions. Better scores are obtained on the worst dives as probability mass is placed on a wide range of outcomes. This behaviour is undesirable, as it leads to uninformative predictions that are also detrimental to the downstream planning. The scoring function $J(\Theta)$ is therefore defined as:

\begin{multline*}
J(\Theta)= \sum_{i=1}^N{\log \hat{p}^{(i)}_\Theta (\mathbf{p}^{(i)}_\text{true}) +
\lambda_{\text{dur}}
\sum_{i=1}^N{\log \widehat{\Delta\tau}^{(i)}_\Theta(\Delta\tau^{(i)}_\text{true})}}
\\
- \lambda_{\text{reg}} \sum_{i=1}^N{ \sum_{j=1}^M{ (\mathbf{p}^{(i, j)} - \overline{\mathbf{p}^{(i,j)}})^2  } }
,
\end{multline*}

where the first term corresponds to the log-likelihood $\ell(\mathcal{D}; \Theta)$ of surfacing locations, the second term corresponds to the duration score with a configurable weight $\lambda_{\text{dur}}$, and the third term penalises the spread of samples generated for each $\mathbf{x}^{(i)}$ with a configurable weight $\lambda_{\text{reg}}$.

We seek to find the optimal parameters $\Theta^* = \argmax_{\Theta}(J(\Theta))$. This high-dimensional optimisation problem has a scoring function that is expensive to evaluate and cannot be computed in closed form. As such, Bayesian optimization is a suitable framework. We opt for BoTorch~\cite{balandat2020botorch,olson2025ax}, whose acquisition function navigates the exploration-exploitation tradeoff to generate promising candidate parameter sets $\Theta_k$ at each iteration $k$ of a total $n_\text{SIM-BO}$ iterations.  

\subsection{Sample-Based Online Planner}\label{subsec:planner}

\begin{algorithm}[!t]
\caption{Sample-Based Online Planner}
\label{alg:planner}
\begin{algorithmic}[1]

\STATE \textbf{Function} \texttt{PlanNextDive}($I$):
\STATE \quad \texttt{allActions} $\leftarrow [\ ]$
\STATE \quad \textbf{for} each $j$ out of $n_\text{threads}$ \textbf{in parallel}
\STATE \qquad \texttt{seed} $\leftarrow $ \texttt{SampleRandomState}($j$)
\STATE \qquad \texttt{action} $\leftarrow $ \texttt{RunUCT}($I$, \texttt{seed})
\STATE \qquad \texttt{allActions.Append(action)}
\STATE \quad $\mathbf{a} \leftarrow$ \texttt{MajorityVote(allActions)}
\STATE \quad \textbf{return} $\mathbf{a}$
\vspace{1mm}
\STATE \textbf{Function} \texttt{RunUCT}($I$, \texttt{randomState}):
\STATE \quad $(\mathbf{p}_0, \tau_0, \mathbf{p}_\text{goal}, \rho, \mathbf{q}(\cdot), \mathbf{b}(\cdot)) \leftarrow\ I$ 
\STATE \quad create root node $v_\text{root}$ from $s_0=(\mathbf{p}_0, \tau_0)$
\STATE \quad \textbf{for} $i$ in $[1, n_\text{trials}]$ \textbf{do}
\STATE \qquad $v_\text{leaf} \leftarrow $ \texttt{TreePolicy}($v_\text{root}$)
\STATE \qquad $\mathbf{s} \leftarrow $ \texttt{GetState}($v_\text{leaf}$)
\STATE \qquad \textbf{if} $\mathbf{s} \in \mathcal{G}$ \textbf{then}
\STATE \qquad \quad $r \leftarrow $ \texttt{SumCosts}($\mathbf{s}_0,\mathbf{s}$)
\STATE \qquad \textbf{else}
\STATE \qquad \quad $r \leftarrow \epsilon_\text{heur} * (d(\mathbf{p}, \mathbf{p}_\text{goal}) / s)$ 
\STATE \qquad \texttt{Backup}($v_\text{leaf}, r$)
\STATE \quad \texttt{action} $\leftarrow $ \texttt{RobustChild}($v_\text{root}$)
\STATE \quad \textbf{return} \texttt{action}
\end{algorithmic}
\end{algorithm}

Gliders surface every few hours and await instructions for the next dive while uploading scientific data. Thus, an \textit{online} planning approach is natural for solving the MDP presented in Section~\ref{subsec:mdp_formulation}, leveraging the time on the surface to derive a navigation plan. At each surfacing, the problem instance $I = (\mathbf{p}_0, \tau_0, \mathbf{p}_\text{goal}, \rho, \mathbf{q}(\cdot), \mathbf{b}(\cdot))$ is assembled using the position $\mathbf{p}_0$ from the glider's GPS fix, time $\tau_0$, goal specification, current forecasts, and bathymetry. 

To perform sample-based online planning and solve the MDP approximately, we leverage Monte Carlo Tree Search (MCTS), an online search algorithm for determining the best action to apply at the current state of an MDP. As a full review of MCTS is out of scope, we refer the interested reader to~\cite{browneSurvey2012} for more details. Specifically, we opt for the Upper Confidence Bounds for Trees (UCT)~\cite{kocsisBandit2006} variant, which treats action selection at each node as an independent multi-armed bandit problem, applying the principles behind the Upper Confidence Bounds (UCB)~\cite{auerFinite2002} algorithm to the tree search setting. Planning is carried out while the glider is at the surface using $n_\text{trials}$ search tree traversals. $n_\text{trials}$ is a parameter that is set on a per-deployment basis to account for the expected time duration of the glider at the surface and the computational resources available.

Pseudocode for the planner is given in Algorithm~\ref{alg:planner}. We use the following enhancements over the basic MCTS scheme. Our planner leverages \textit{double progressive widening}, which is suitable for scenarios with stochastic transitions and continuous state spaces~\cite{chaslotProgressive2008,couetouxContinuousUpperConfidence2011}. It is able to achieve a suitable compromise between bias (induced by the first expanded nodes) and variance (in the result of trials for each state). We note that, despite the discretisation of the actions, we opt for progressive widening in the action space also. This is done to keep the branching factor low and increase the tree depth, enabling us to plan over a longer horizon. 

Furthermore, we use \textit{root parallelism} in order to take advantage of multi-core architectures despite the sequential nature of MCTS. This creates $n_\text{threads}$ search trees that progress independently from the given root state. The final results are aggregated by majority voting to obtain a single action~\cite{cazenaveParallelizationUCT2007,chaslotParallelMonteCarloTree2008}. This improves the robustness of the navigation plan to both noise in the current forecasts and action choices early in the tree search.

We also highlight the following differences from a standard implementation:

\begin{itemize}
	\item \textbf{Use of heuristic.} We do not perform rollouts to estimate the value of leaf nodes as sampling transitions requires running a computationally intensive simulator. This is impractical given the time budget available at each surfacing. Instead, we use a heuristic (line 18) that estimates the time to reach the goal based on the distance to goal and constant speed. The error term $\epsilon_\text{heur}$ of the heuristic was calibrated based on preliminary deployment data and was set to approximately $1.77$. Equivalently, the typical trajectory length was $1.77\times$ that of the straight line.
	 \item \textbf{Prioritising STG action.} When expanding a node, we always add the child corresponding to the STG action ($\alpha = 0$) first. Further actions are sampled randomly from $\mathcal{A}$ as visit counts increase and the progressive widening condition is triggered. This prioritises an action that leads to good returns in most situations and corresponds to the standard navigation baseline.
\end{itemize}

\section{\MakeUppercase{\syslong}}

In this section, we describe the components required for translating our methodological and algorithmic contributions discussed in the previous section to a field-validated system that performs long-term autonomous navigation planning for underwater gliders. The \textit{\syslong} was designed and implemented in collaboration with operational stakeholders in order to enhance operational practicality and safety.

\subsection{Automated Goal Management}~\label{sub:automated_goal_management}

The previous section specified problem instances $I$ with a single goal $\mathbf{p}_\text{goal}$. However, glider missions typically consist of many such goals. A common use case, which we target, is to have the glider sample data between a set of predefined points, i.e., a \textit{transect} in the case of two points, and the corners of a polygon more generally. We therefore assume we are given an ordered list of points $\{\mathbf{p}^{(i)}_{\text{goal}}\}_{i=1}^{n_{\text{goals}}}$. The system advances to the next goal in the list upon achieving the current one, and cycles to the beginning of the list once the end is reached. 

For determining whether the goal was achieved, i.e., whether $\mathbf{s} \in \mathcal{G}$, an additional condition must be checked in \sysshort besides surfacing within the goal region. A goal may also be detected as having been achieved while underwater using dead reckoning (which we note may be inaccurate). This reflects the operation of the on-board glider software.

\begin{figure}[t]
\begin{center}
  \includegraphics[width=\columnwidth]{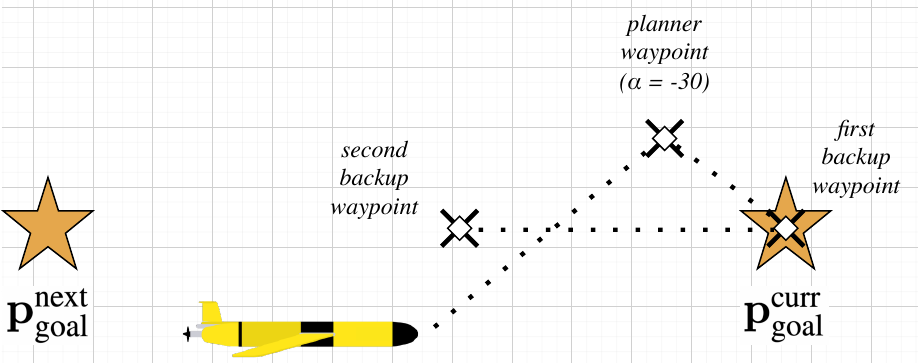}
 \caption{Visualisation of waypoint list generation. The first waypoint corresponds to the planner action (here, $\alpha=-30$). The first backup waypoint is set to be the current goal as otherwise the goal would be overshot. The second backup waypoint is aimed straight at the next goal.}
\label{fig:mission_plan}
\end{center}
\end{figure}

\begin{figure*}[t]
\begin{center}
  \includegraphics[width=0.9\textwidth]{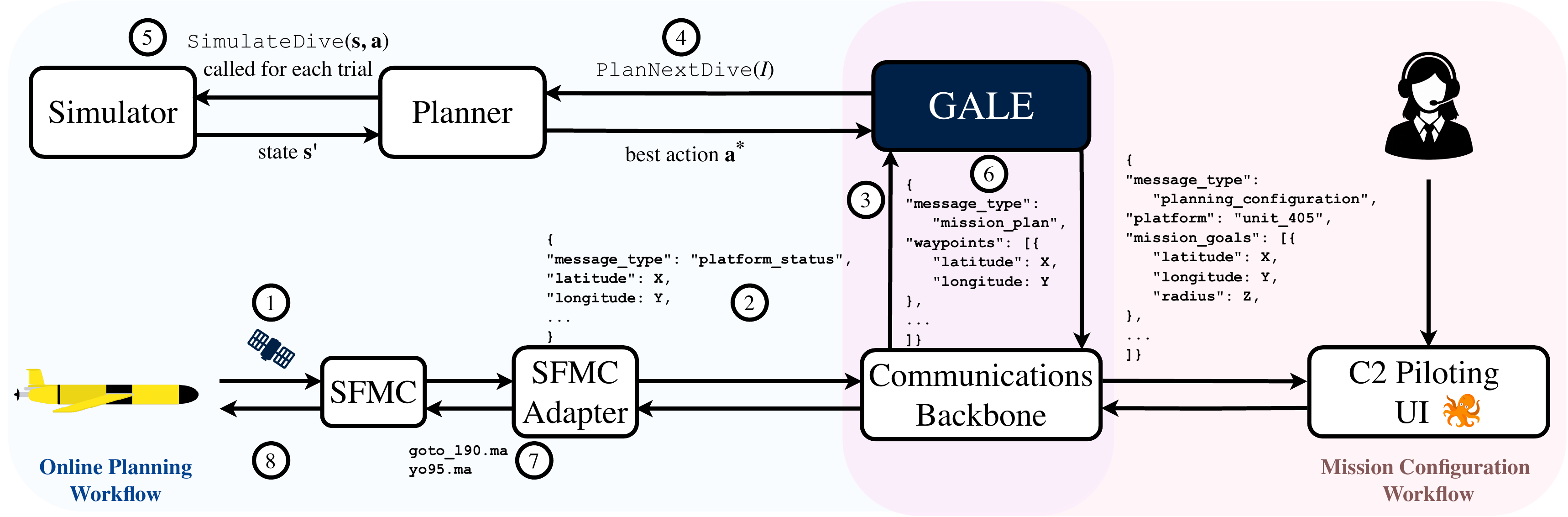}
 \caption{Diagram illustrating the components and steps of the two system workflows. The Mission Configuration Workflow (right) is executed once at the start the mission, while the Online Planning Workflow (left) is executed every few hours upon the surfacing of the glider.  
}
\label{fig:flow_diagram}
\end{center}
\end{figure*}
\subsection{Translation of Control Instructions}~\label{sub:waypoint_list}

Actions $\mathbf{a} = (\alpha, \mathbf{c})$ output by the planner must be translated to \textit{files} that can be understood by a Slocum glider. Two types of files are required: a \textit{waypoint list} specifying an ordered sequence of coordinates that the glider should reach, and an \textit{instruction set} containing the control parameters $\mathbf{c}$.

To generate the waypoint list, we use the $\alpha$ component of the planner action for computing the first waypoint and several backup waypoints derived using straight-to-goal bearings. Since waypoint lists are recomputed and updated at each surfacing, the vehicle can steer according to planner output during normal operation and use a sensible backup trajectory otherwise. As mentioned in Section~\ref{sec:gliders}, several failures in transferring control instructions can occur in a row. Therefore, waypoint lists must extend well beyond a single dive so that the vehicle can continue its mission uninterrupted in the case of communication failures.

A waypoint list generation example is shown in Figure~\ref{fig:mission_plan}. The algorithm, which we describe at a high level due to space limitations, takes as input the action $\alpha$, glider position $\mathbf{p}_0$, the position of the current goal $\mathbf{p}_\text{goal}^{\text{curr}}$, the position of the next goal $\mathbf{p}_\text{goal}^{\text{next}}$, the goal tolerance radius $\rho$, the waypoint distance $\rho_\text{wpt}$, and number of backup waypoints $n_\text{bck}$. The algorithm first determines the waypoint corresponding to the planner action. If distance to the goal is low, the goal itself is used as the waypoint to avoid oscillating around it. Otherwise, the waypoint is set along the heading of the planner action at the specified waypoint distance. Backup waypoints towards the current goal and, if required, the next goal are added. The bearings for the backup waypoints are computed starting from the previous waypoint. Additional backup waypoints towards a particular goal are not necessary if the previous waypoint is already within the same goal region. If a backup waypoint overshoots the goal region, it is replaced with the goal itself.

Generating the instruction set file involves a format translation of the control parameters $\mathbf{c}$. Following the guidance of glider pilots, \sysshort supports specifying a number of fixed instruction sets $\mathcal{K}$ from which $\mathbf{c}$ is chosen when the mission is initialised. This ensures the control parameters are within a field-tested range of configurations, protecting the assets and ensuring pilot trust in the system. As previously mentioned, for simplicity, we assume that a fixed $\mathbf{c}$ is used for the duration of a particular mission. This assumption is realistic for the deployments considered in our evaluation, which require sampling of scientific data at regular time and distance intervals. Dynamically varying the instruction set on a per-action basis is left for future work.

\subsection{System Components and Integration}

The \sysshort system implements the methodology described in Section~\ref{sec:methods} and is a service deployed to an on-premise server. It is capable of controlling several gliders simultaneously and fetches the most recent current forecasts daily. In this section, we describe the components that are required besides \sysshort to achieve the full control loop of the glider.

\noindent \textbf{C2 Communications Backbone}. This open-source component~\cite{harrisOceanidsC2Integrated2020,mars2025backbone} developed by the UK National Oceanography Centre (NOC) enables real-time transfer of vehicle information and control instructions using standardised APIs~\cite{mars2025message}.

\noindent \textbf{SFMC}. Slocum Fleet Mission Control (SFMC) provides interfaces for controlling gliders manually or programmatically. It enables transferring files containing waypoints and control parameters by placing them in a dedicated directory from which they are transmitted via an Iridium satellite link.

\noindent \textbf{C2 Piloting User Interface}. This web-based interface supports configuring missions for, monitoring, and piloting marine autonomous systems. It is complementary to SFMC and unifies piloting interfaces for several platforms~\cite{harrisOceanidsC2Integrated2020}. 

\subsection{Workflows}~\label{subsec:workflows}

Two workflows are illustrated in Figure~\ref{fig:flow_diagram}.

\noindent \textbf {Mission Configuration Workflow}. This workflow is executed once at the start of a mission. It involves the use of the C2 Piloting UI to indicate the glider, the list of goals $\{ \mathbf{p}_\text{goal}^{(i)} \}$, and permitted instruction sets $\mathcal{K}$. The \sysshort system initialises the tracking information in persistent storage. From this point onwards, \sysshort will track the mission goals and respond to post-surfacing updates from the glider with navigation plans.

\noindent \textbf {Online Planning Workflow}. This workflow achieves the full glider control loop. Upon surfacing, the glider calls in to SFMC (1). Over satellite, it relays status information (time, GPS fix) and transfers scientific data while awaiting instructions from \sysshort, which creates a corresponding problem instance $I$ (2, 3). The Planner (Algorithm~\ref{alg:planner}) is invoked using $I$, which calls the Simulator (Algorithm~\ref{alg:glidersim}) on the order of $10^3-10^4$ times in each thread (4,5). The best action $\mathbf{a}$ is determined by the Planner and converted to a waypoint list and instruction set, which are published to the Backbone (6). The messages are placed as files in the appropriate directory on SFMC (7). They are transmitted to the glider and are used to execute the next dive (8). 

It is worth reiterating that steps 1 and 8, which involve communication over satellite, are not fully reliable. In preliminary deployments, the rate of successful transfer of waypoint lists and instruction sets was approximately 89\%, with further significant drops during periods with adverse weather conditions. Since navigation plans cover a longer horizon than individual dives (on the order of a day rather than hours), the glider was able to continue its mission.

\subsection{System Configuration}

Configuring the system for a new deployment requires the following steps:

\begin{itemize}
  \item The scientific stakeholders supply the mission goals $\{ \mathbf{p}_\text{goal}^{(i)} \}$ and the goal tolerance radius $\rho$.
  \item The glider pilots specify the permitted instruction sets $\mathcal{K}$ that are appropriate for the chosen vehicle and the scientific objectives.
  \item Relevant data from historical glider deployments is identified and retrieved (e.g., from public repositories~\cite{noc2025bodc}). Alignment to the upcoming deployment in terms of location, vehicle type, and time period are important for the faithfulness of the simulator.
  \item The GALE system administrator performs simulator parameter tuning (Section~\ref{subsec:sim_tuning}) using the historical data. Planner parameter tuning is also performed given the tuned parameters for the simulator. An instance of the GALE system is configured and deployed to computational infrastructure.
  \item The vehicle is deployed and piloted manually towards the operational area of the campaign.
  \item The pilots execute the Mission Configuration Workflow (Section~\ref{subsec:workflows}). This signals the start of the automated piloting, completing the setup for this deployment.
\end{itemize}

\section{EXPERIMENTAL SETUP}

\subsection{Field Experiments}\label{sub:deployments}

\begin{figure}[t]
\begin{center}
  \includegraphics[width=\columnwidth]{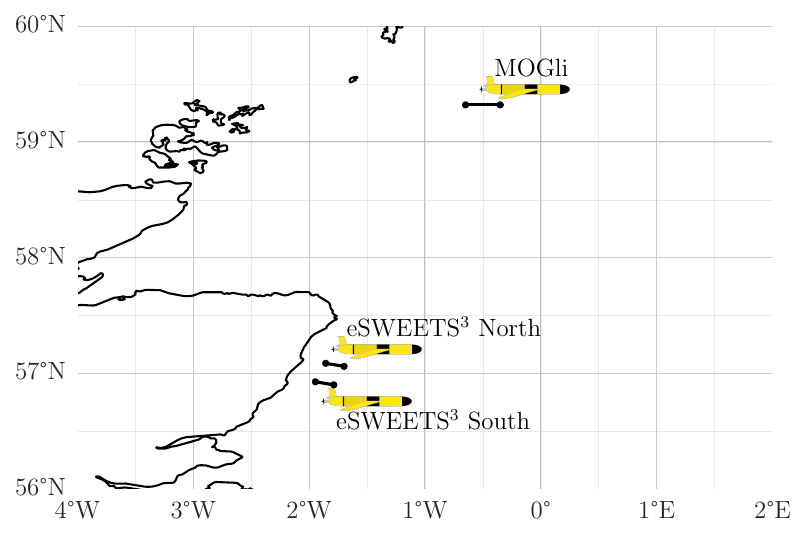}
 \caption{Field deployment locations in the North Sea.}
\label{fig:deployment-map}
\end{center}
\end{figure}

Our system was evaluated in two observational campaigns in the North Sea: \textit{MOGli} (one glider) and \textit{\Esw} (two gliders simultaneously). In these deployments, gliders were operated by NOC on behalf of external partners the UK Met Office and the University of East Anglia. A transect formed of two goal positions was used in all deployments. Deployment coordinates and summary statistics are given in Tables~\ref{tab:deployment-goals} and \ref{tab:deployment-stats}. A map of the three field deployments is shown in Figure~\ref{fig:deployment-map}.

\begin{table}[!t]
\centering
\caption{Deployment goal locations and parameters. All distances are given in km.}
\label{tab:deployment-goals}
\begin{tabular}{l|c|c|c|c|c|c}
\hline
\textbf{Deployment} & \textbf{Goal 1} & \textbf{Goal 2} & \textbf{Dist.} & $\rho$ & $\rho_\text{wpt}$ & $n_\text{bck}$ \\
\hline
MOGli &
\begin{tabular}[c]{@{}c@{}}
$59.32^\circ$N\\
$0.35^\circ$W
\end{tabular} &
\begin{tabular}[c]{@{}c@{}}
$59.32^\circ$N\\
$0.65^\circ$W
\end{tabular} &
17.0 &
2.0 &
7.0 & 
2
\\
\hline
\Esw~N &
\begin{tabular}[c]{@{}c@{}}
$57.089^\circ$N\\
$1.856^\circ$W
\end{tabular} &
\begin{tabular}[c]{@{}c@{}}
$57.062^\circ$N\\
$1.698^\circ$W
\end{tabular} &
10.0 &
1.0 &
7.0 & 
2
\\
\hline
\Esw~S &
\begin{tabular}[c]{@{}c@{}}
$56.929^\circ$N\\
$1.945^\circ$W
\end{tabular} &
\begin{tabular}[c]{@{}c@{}}
$56.903^\circ$N\\
$1.787^\circ$W
\end{tabular} &
10.0 & 
1.0 &
7.0 & 
2
\\
\hline
\end{tabular}
\end{table}

\begin{table}[!t]
\centering
\caption{Deployment summary statistics.}
\label{tab:deployment-stats}
\begin{tabular}{l|c|c|c|c}
\hline
\textbf{Deployment} &
\shortstack{\textbf{Transects}\\\textbf{completed}} &
\shortstack{\textbf{Duration}\\\textbf{(days)}} &
\shortstack{\textbf{Total}\\\textbf{dives}} &
\shortstack{\textbf{Distance}\\\textbf{travelled (km)}} \\
\hline
MOGli          & 41 & 42 & 343 & 665 \\
\Esw~North  & 20 & 27 & 150 & 173  \\
\Esw~South  & 23 & 22 & 147 & 207  \\
\hline
\end{tabular}
\end{table}

\noindent \textbf{MOGli.} \textit{Met Office Glider (MOGli)} is an ongoing joint project between the UK Met Office (MO) and NOC. Measurements of ocean properties are gathered by a single glider and transmitted periodically to the MO. This data is used to refine MO forecast models. The use of \sysshort to pilot the glider leads to a virtuous circle as AMM15 forecasts are used to derive navigation plans, while the scientific data itself is used to improve the forecasts.

The glider was deployed in a region with low bottom temperature variance since this was part of the MOGli scientific objectives. Fully autonomous control operated from 2 May to 3 July 2025, following preliminary experiments that began in November 2024. A pause in autonomous control was in place from 21 May to 7 June 2025 while the glider was recovered, recharged, and redeployed to the region. 

The action space was chosen to contain relative bearings $\alpha \in \{-40, -20, 0, 20, 40\}$. For the fixed control parameters $\mathbf{c}$, a value of $n_\text{yos}=5$ yos was chosen to lead to an operationally appropriate dive duration of approximately $3$ hours. A target depth of $z_\text{bottom}=150$ was used, which exceeds the depth of the seabed in this region. This was driven by the requirement to sample the entire water column, and hence the glider uses the altimeter to inflect when close to the seabed, a feature also supported by our simulator.

\noindent \textbf{\Esw.} In the \textit{enabling Sustainable Wind Energy Expansion in Seasonally Stratified Seas (\Esw)} project, two Slocum gliders were deployed close to the Kincardine floating offshore wind farm. The gliders were positioned upstream (referred to as \Esw~North) and downstream (\Esw~South) of the wind farm to assess the impacts of the floating structures on the local turbulent mixing, which in turn influences oxygen availability, nutrient supplies, and biological productivity. 

The project deployments were performed in two phases. In the first phase (May - August 2025), gliders were piloted manually using standard procedures. In the second phase (August - October 2025), \sysshort was used to control the gliders following preliminary experiments. Fully autonomous operation was in place from 2 September to 30 October 2025 for \Esw~North and from 18 September to 30 October 2025 for \Esw~South.

The chosen action space contained relative bearings $\alpha \in \{-90, -60, -30, 0, 30, 60, 90\}$, a wider range than in MOGli to account for the effect of strong tidal currents, which were aligned roughly orthogonal to the glider transects. $n_\text{yos}=10$ yos were used, leading to dives lasting approximately $4$ hours in this shallower region. A target depth of $z_\text{bottom}=150$ was used which, as in MOGli, exceeds the bathymetric limit.

The \Esw~deployments were subject to additional operational requirements. As the gliders were deployed in areas with strong tidal currents, there was a risk of vehicle loss if it drifted away from the mission area, especially into the wind farm or shipping lanes. \textit{Safety polygons} were defined, and additional functionality was implemented such that, if the glider drifted outside of the safety polygon, the algorithm described in Section~\ref{sub:waypoint_list} is replaced with a simplified navigation plan aiming directly for the center of the polygon. Normal navigation was resumed upon re-entry. Furthermore, as continuing to sample along the transect upon re-entry was deemed more important for the scientific objectives than turning back to achieve the goal, the system switched to the next goal if the glider had already progressed past 90\% of the transect before exiting the safety polygon.

\textbf{Baseline.} As a baseline for the Planner, we consider the \textit{Straight-to-Goal} approach, which always selects the action with $\alpha=0$. For an equitable comparison, the glider remains under the autonomous control of \sysshort when Straight-to-Goal is used. Since only one of the two methods can control the glider at any one time, an alternating-control approach was taken. The two methods took turns to control the glider after the achievement of $2$ goals (equivalently, the algorithm was switched after every back-and-forth transect traversal). Even though the currents experienced by the two algorithms differ, we expect that any bias thus introduced is removed in expectation.

For MOGli, alternating-control was followed throughout. For \Esw, exclusive use of Straight-to-Goal was mandated by operational concerns until 18 September for \Esw{} North and 2 October for \Esw{} South. \sysshort was subsequently set to issue Planner waypoints for a comparable duration (until 14 October for \Esw{} North and 15 October for \Esw{} South). For the final period, the algorithms were switched to alternating-control after every back-and-forth transect traversal as in MOGli.

\subsection{Simulation Experiments}

We present a set of experiments in simulation whose setup mirrors the deployments listed above. They are an idealised version of the real-world experiments, wherein file transfers always succeed and currents experienced are well-represented by the forecasts. Underwater goal achievement detection (Section~\ref{sub:automated_goal_management}) and the additional functionality for \Esw{} (Section~\ref{sub:deployments}) are not supported in simulation. The simulator and planner parameters are otherwise set identically to those used in the field deployments. 

We select three scenarios for each deployment to be replayed in simulation out of the tens of transects completed in the real world (see counts in Table~\ref{tab:deployment-stats}). A scenario consists of a starting state (glider position and time) and goal assigned to the glider, which implies a set of historical current forecasts corresponding to the starting state. The same scenarios can be ran as many times as desired with different realisations of the uncertainty, allowing statistically robust comparisons. The baseline remains the same (Straight-to-Goal). In contrast with the field experiments, in simulation the two methods can be compared using identical conditions. 

To determine the scenarios, we first assess the number of dives required by Straight-to-Goal to complete each transect averaged over $200$ repetitions. We pick the starting conditions for the transects with the minimum (\textit{Favourable}), median (\textit{Neutral}), and maximum (\textit{Unfavourable}) dives required on average for the baseline to reach the goal. This yields scenarios that represent a range of conditions experienced by the glider during the deployment periods. We then re-run both the Planner and Straight-to-Goal over $1000$ random seeds to report the final results.

\subsection{Metrics}

\noindent \textbf{Planner.} To evaluate the planner, we report the total transect duration (i.e., the cumulative cost as defined in Section~\ref{subsec:mdp_formulation}), which is the objective that the method optimises for. As secondary metrics, we report the number of dives (highly correlated with total duration) and the total distance covered for each traversal. All ± entries in tables denote 95\% confidence intervals. Occasionally, navigation plans were not generated or followed by the glider due to planned (e.g., software upgrades) or unplanned (e.g., anomalies causing abort of the dive) concerns. Even though these events typically impacted a single surfacing, they compromise the transect validity. Such transects were excluded from the planner evaluation.

\noindent \textbf{Simulator.} To evaluate the simulator, we report the true and simulated dive distances and durations as well as their differences (deltas). Since datasets contain different numbers of dives, we report the mean (per-dive) log-likelihood $\bar{\ell}$ instead of the summed log-likelihood $\ell$. These metrics are assessed on the dives recorded during the live deployment, which were not used for simulator optimisation or validation. Furthermore, as the simulator is stochastic and produces several plausible next states, we report statistics with respect to the sample means obtained by drawing $40$ samples for each dive. 

\subsection{Simulator and Planner Parameter Optimisation}

To optimise the simulator parameters (Section~\ref{subsec:sim_tuning}), we use historical glider data and current forecasts gathered in preliminary deployments (concretely, February - March 2025 for MOGli and May - June for \Esw{}). The dataset for simulator parameter optimisation is pooled across the two \Esw{} gliders given the proximity of the two regions. Therefore, the same simulator parameters are used in both \Esw{} North and South. We use AMM15 forecasts archived by our system for simulator parameter optimisation. However, such data can also be obtained through publicly available oceanographic data repositories (e.g.,~\cite{noc2025bodc} and~\cite{copernicus2025marine}).

The dives are split randomly into training and validation sets. Parameters are optimised using Bayesian optimisation (BO) on the training set, and the best parameters on the validation set with respect to $J(\Theta)$ are selected. We apply several data cleaning steps by excluding dives that did not terminate due to normal surfacing, those with speed $2$ standard deviations outside the mean, and with a difference between heading and surface location greater than a $75$ degree threshold for MOGli and $80$ for \Esw{}. These criteria resulted in excluding $19.58\%$ of dives for MOGLi and $5.30\%$ dives for \Esw{}. $n_\text{SIM-BO}=500$ iterations of BO were used as the scoring function plateaued before this step in all experiments.

In terms of planner parameters, we use $n_{\text{threads}}=8$ parallel search trees for both deployments. We set $n_\text{trials}=10000$ and $n_\text{trials}=5000$ for MOGli and \Esw{} respectively, as for the latter deployment less data is transmitted and therefore less online planning time is available. We found that the planner hyperparameters (exploration constant, progressive widening parameters, maximum tree depth) are sensitive to simulator parameters. Therefore, after fixing simulator parameters, planner hyperparameters are optimised downstream with BO under the objective of minimising the cumulative cost to reach the goal.

\section{RESULTS}

\begin{figure}[t]
\begin{center}
  \includegraphics[width=0.85\columnwidth]{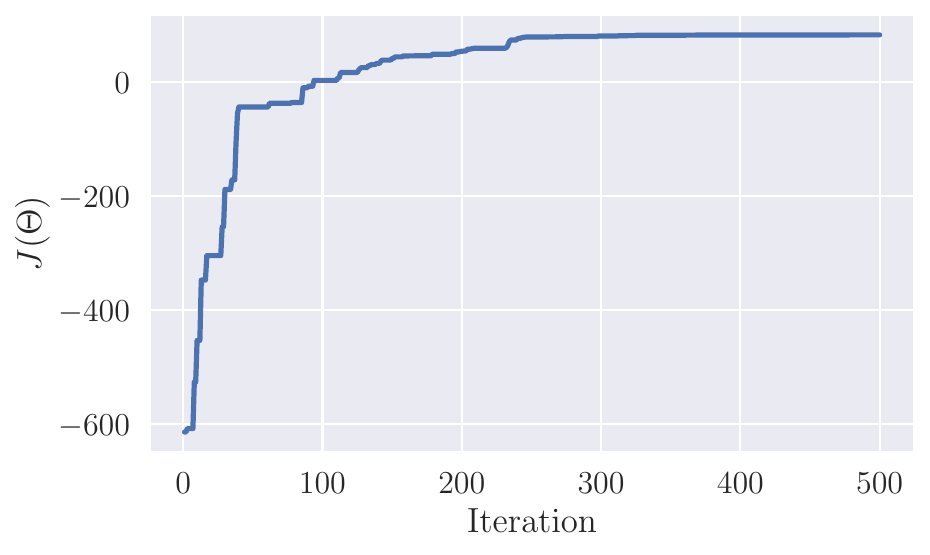}
 \caption{Learning curve showing the best value of $J(\Theta)$ found as a function of Bayesian optimisation iterations using the MOGli training data.}
\label{fig:simulator_optimisation}
\end{center}
\end{figure}

\begin{table}[!t]
\centering
\caption{Comparison of simulated and true dive metrics averaged over all dives in the real-world deployments.}
\label{tab:dive_metrics}
\begin{tabular}{cccc}
\hline
\textbf{Metric} &
\textbf{MOGli} &
\textbf{\Esw~North} &
\textbf{\Esw~South} \\
\hline
\shortstack[c]{True dive \\ distance (m)} &
2733 & 5650 & 4679 \\
\hline
\shortstack[c]{Simulated dive \\ distance (m)} &
2119 & 4991 & 4362 \\
\hline
\shortstack[c]{Distance \\ delta (m)} &
1448 & 1555 & 1080 \\
\hline
\shortstack[c]{True dive \\ duration (s)} &
10421 & 15923 & 12935 \\
\hline
\shortstack[c]{Duration \\ delta (s)} &
345 & 1481 & 1377 \\
\hline
\shortstack[c]{Mean \\ log-likelihood $\bar{\ell}$} &
-9.29 & -78.67 & -133.90 \\
\hline
\end{tabular}
\end{table}

\subsection{Simulator Evaluation}\label{sub:simulator_evaluation}

\begin{figure*}[t]
	\begin{center}
		\includegraphics[width=0.95\textwidth]{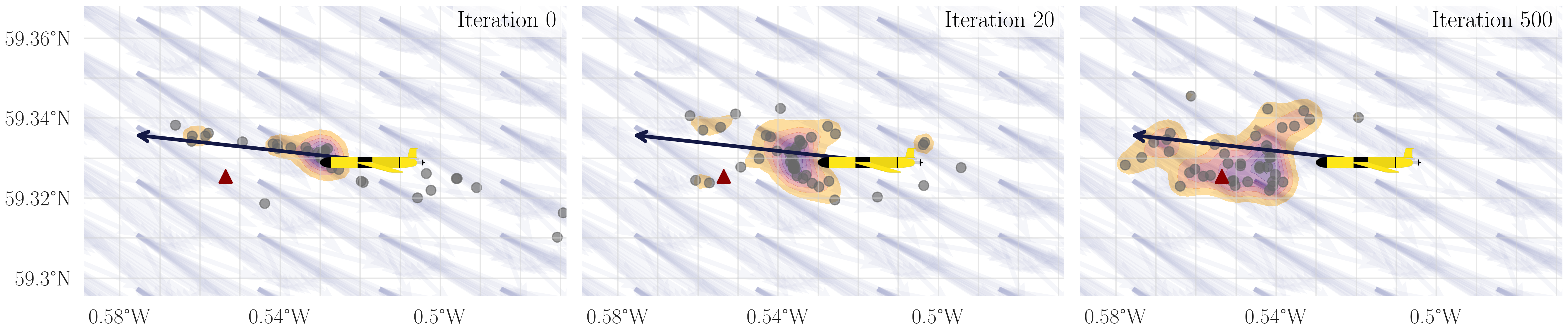}
		\caption{Progress of the proposed simulator parameter optimisation procedure (Section~\ref{subsec:sim_tuning}) on one of the dives in the MOGli deployment. Simulator parameters are progressively adjusted such that the surfacing locations generated by the simulator (grey circles) best match the true surfacing location (red triangle).
		}
		\label{fig:three_simulator_illustration}
	\end{center}
\end{figure*}

Figure~\ref{fig:simulator_optimisation} shows the progress of the scoring function $J(\Theta)$ during the steps of the Bayesian optimisation procedure on the MOGli data. Additionally, in Figure~\ref{fig:three_simulator_illustration}, we show snapshots of simulator outputs for a MOGli dive at the beginning, intermediate, and final steps of the procedure. The refinement of the simulator output distribution with respect to the true surfacing location is evident. 

Several values for the spread penalty $\lambda_{\text{reg}}$ were tested, and the parameters that obtained the best $J(\Theta)$ on the validation set were chosen ($\lambda_{\text{reg}} = 1$ for MOGli, corresponding to the figures shown, and $\lambda_{\text{reg}} = 0$ for \Esw{}). The dive durations can realistically be treated as approximately constant in the MOGli region given the assumption of fixed control parameters and the fact that the bathymetry is consistent throughout the transect. Therefore, the duration score coefficient for MOGli is $\lambda_{\text{dur}}=0$. In contrast, we set $\lambda_{\text{dur}}=1$ in the \Esw{} region as the eastern ends of the \Esw{} transects are $20-40$ m deeper than the western ends. The variation in bathymetry induces variation in dive durations given fixed dive control parameters (i.e., dive durations increase in deeper regions). The standard deviation $\sigma=700$ was estimated based on data from the first phase which used manual piloting (Section~\ref{sub:deployments}).

In Table~\ref{tab:dive_metrics}, we report several metrics concerning simulator performance on the deployment data (not seen during training or model selection). The distances of the dives output by the simulator were close to their true average on \Esw{}, but noticeably underestimated for MOGli. The distance between the true and simulated surfacing locations is around $1.5$ km on average for all deployments (though it is inherently challenging to draw conclusions by comparing an individual sample with the empirical mean of the simulator distribution). In contrast, the dive durations were more accurate for MOGli given the aforementioned variation in bathymetry for \Esw{}. The mean log-likelihood $\bar{\ell}$ degraded on the deployment data compared to the training data, as would be expected. On the deployment data, the optimised simulator parameters yielded scores of $\bar{\ell}=-9.29$ for MOGli and $\bar{\ell}=-106.00$ for \Esw{}, which are significantly better than those of randomly chosen parameters at the start of optimisation ($\bar{\ell} \approx -32$ and $\bar{\ell} \approx -285$ respectively). This evidences the benefits of the proposed tuning procedure. The log-likelihood scores are worse for \Esw{} compared to MOGli due to the stronger currents in this tidal region.

We note that there are many possible causes for predictive errors in the simulator, which include the following: forecasts being wrong, the steady-state approximation of our simulator, mismatches between the physical glider used for simulator optimisation and the one used in the real world (distinct vehicles with variations in payload and ballasting), and noise in how the glider executes the control commands. It is highly likely that all these factors are at play and, while simulator accuracy can be improved, some degree of error is unavoidable.

\subsection{Planner Evaluation in Simulation}

\begin{table}[!t]
\centering
\caption{Results comparing the Straight-to-Goal and Planner algorithms across several scenarios in simulation. Bold indicates the better-performing algorithm.}
\label{tab:simulation-results}
\resizebox{1\columnwidth}{!}{
\begin{tabular}{lllccc}
\hline
\textbf{Deployment} & \textbf{Scenario} &
\textbf{Algorithm} & 
\shortstack{\textbf{Mean}\\\textbf{Duration}\\\textbf{(hrs)}} &
\shortstack{\textbf{Mean}\\\textbf{Number}\\\textbf{Dives}} &
\shortstack{\textbf{Mean}\\\textbf{Path}\\\textbf{Length}\\\textbf{(km)}} \\
\hline

MOGli & Fav. & STG & 22.48{\tiny$\pm$0.30} & 7.68{\tiny$\pm$0.10} & 17.31{\tiny$\pm$0.05} \\

& & Planner & \textbf{22.13}{\tiny$\pm$0.30} & \textbf{7.57}{\tiny$\pm$0.10} & \textbf{16.79}{\tiny$\pm$0.06} \\

& Neutral & STG & 22.93{\tiny$\pm$0.31} & 7.87{\tiny$\pm$0.11} & 17.66{\tiny$\pm$0.05} \\

& & Planner & \textbf{22.38}{\tiny$\pm$0.30} & \textbf{7.68}{\tiny$\pm$0.10} & \textbf{16.87}{\tiny$\pm$0.06} \\

& Unfav. & STG & 23.60{\tiny$\pm$0.34} & 8.10{\tiny$\pm$0.12} & 17.85{\tiny$\pm$0.06} \\

& & Planner & \textbf{22.84}{\tiny$\pm$0.32} & \textbf{7.80}{\tiny$\pm$0.11} & \textbf{16.92}{\tiny$\pm$0.06} \\

\hline

\Esw & Fav. & STG & 55.67{\tiny$\pm$1.40} & 10.01{\tiny$\pm$0.26} & 52.31{\tiny$\pm$1.54} \\

North & & Planner & \textbf{48.70}{\tiny$\pm$1.44} & \textbf{8.96}{\tiny$\pm$0.28} & \textbf{41.10}{\tiny$\pm$1.45} \\

& Neutral & STG & 65.70{\tiny$\pm$0.87} & 11.33{\tiny$\pm$0.16} & 52.92{\tiny$\pm$1.15} \\

& & Planner & \textbf{62.91}{\tiny$\pm$0.92} & \textbf{10.87}{\tiny$\pm$0.17} & \textbf{46.67}{\tiny$\pm$1.06} \\

& Unfav. & STG & 64.07{\tiny$\pm$1.18} & 11.79{\tiny$\pm$0.22} & 65.20{\tiny$\pm$1.29} \\

& & Planner & \textbf{55.82}{\tiny$\pm$1.43} & \textbf{10.24}{\tiny$\pm$0.27} & \textbf{54.57}{\tiny$\pm$1.40} \\

\hline

\Esw & Fav. & STG & 43.46{\tiny$\pm$1.47} & \textbf{9.70}{\tiny$\pm$0.36} & 43.60{\tiny$\pm$1.58} \\

South & & Planner & \textbf{41.81}{\tiny$\pm$1.50} & 9.78{\tiny$\pm$0.40} & \textbf{40.09}{\tiny$\pm$1.50} \\

& Neutral & STG & 55.63{\tiny$\pm$1.14} & 11.80{\tiny$\pm$0.28} & 53.79{\tiny$\pm$1.23} \\

& & Planner & \textbf{47.29}{\tiny$\pm$0.98} & \textbf{10.08}{\tiny$\pm$0.25} & \textbf{43.09}{\tiny$\pm$0.95} \\

& Unfav. & STG & 61.99{\tiny$\pm$0.91} & 13.04{\tiny$\pm$0.22} & 68.77{\tiny$\pm$1.00} \\

& & Planner & \textbf{57.05}{\tiny$\pm$1.11} & \textbf{12.52}{\tiny$\pm$0.29} & \textbf{59.21}{\tiny$\pm$1.12} \\

\hline
\end{tabular}
}
\end{table}

\begin{figure*}[t]
\begin{center}
  \includegraphics[width=0.92\textwidth]{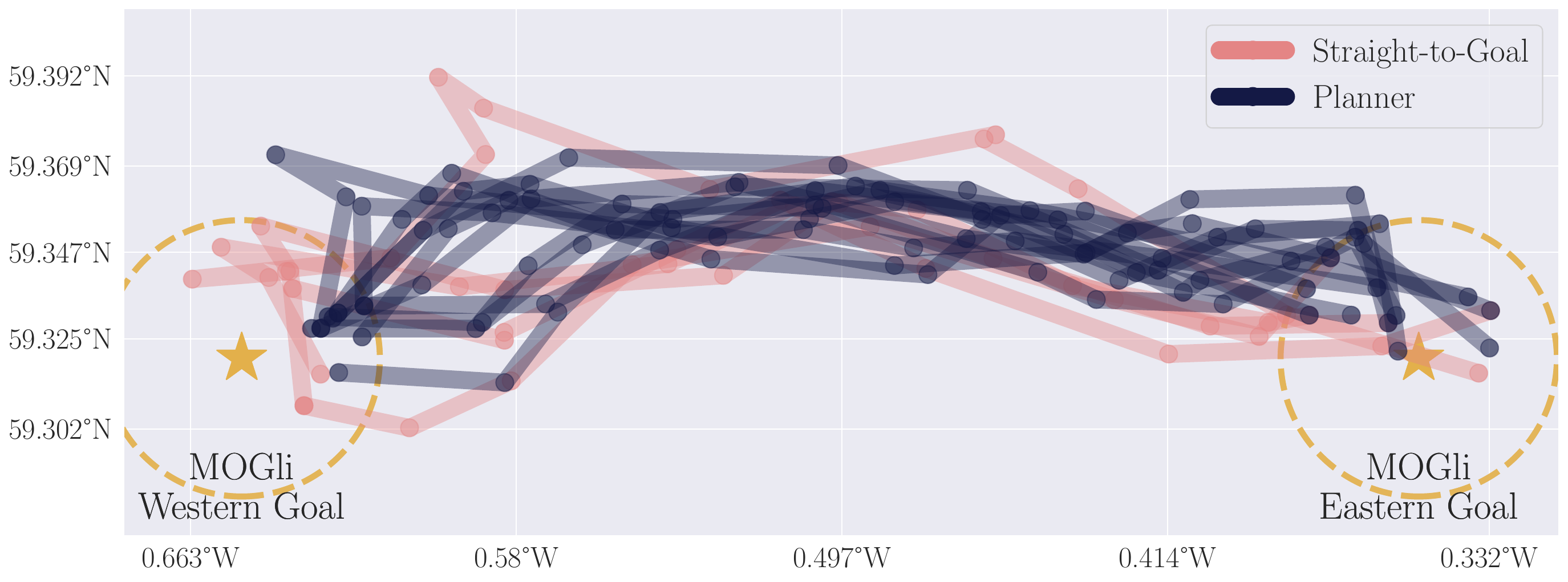}
 \caption{Transects completed by the proposed \sysshort system in the MOGli field deployment using the Planner and the Straight-to-Goal baseline. Only transects in which all transfers of control instructions succeeded are shown. The goal radii are represented by two dotted circles around the goals. The Planner results in transects with shorter minimum, mean, and maximum durations. Transects produced by the two methods are also discernible visually. The transects completed by the Planner have statistically significantly shorter path lengths and avoid overshooting the western goal. 
}
\label{fig:mogli_trajectories}
\end{center}
\end{figure*}

Table~\ref{tab:simulation-results} shows the results comparing the Straight-to-Goal and Planner algorithms in simulation. The Planner improves over Straight-to-Goal in duration and path length in all settings tested. The most significant improvement is obtained for the Favourable scenario in the \Esw{} North deployment, in which the planner reduces the mean transect duration by $7$ hours and the path length by $11$ km. For MOGli, the reductions in duration over the baseline are small in an absolute sense at slightly over $30$ minutes per transect, equating to a $2.39\%$ decrease. This region does not exhibit strong currents, as corroborated by the small differences in the performance of STG between the most Favourable and Unfavourable scenarios. Therefore, our currents-aware approach does not have much room for improvement over the STG algorithm. The improvements are more pronounced for \Esw{}, where the Planner leads to decreases of $9.88\%$ and $8.92\%$ respectively in dive duration. The reductions in path length of $4.23-13.94\%$ are more pronounced than the duration reductions. This is an important secondary metric as sampling along trajectories that are as straight as possible facilitates robust environmental data analysis and helps achieve many scientific objectives.

\subsection{Planner Evaluation in Field Deployments}

\begin{table*}[!t]
\centering
\caption{Results comparing the Straight-to-Goal and Planner algorithms in the field deployments. Bold denotes the better-performing algorithm for each deployment and metric.}
\label{tab:deployment-results}
\begin{tabular}{llccccccc}
\hline
\textbf{Deployment} &
\textbf{Algorithm} &
\shortstack{\textbf{Mean}\\\textbf{Duration}\\\textbf{(hrs)}} &
\shortstack{\textbf{Min}\\\textbf{Duration}\\\textbf{(hrs)}} &
\shortstack{\textbf{Max}\\\textbf{Duration}\\\textbf{(hrs)}} &
\shortstack{\textbf{Mean}\\\textbf{Number}\\\textbf{Dives}} &
\shortstack{\textbf{Min}\\\textbf{Number}\\\textbf{Dives}} &
\shortstack{\textbf{Max}\\\textbf{Number}\\\textbf{Dives}} &
\shortstack{\textbf{Mean}\\\textbf{Path}\\\textbf{Length (km)}} \\
\hline
MOGli & Straight-to-Goal &
23.09 & 16.00 & 31.70 &
8.45 & 6 & 11 &
22.58 \\

      & Planner &
\textbf{22.68} & \textbf{14.35} & \textbf{26.98} &
\textbf{8.28} & \textbf{5} & \textbf{10} &
\textbf{20.52} \\
\hline
\Esw~North & Straight-to-Goal &
40.38 & 19.80 & 83.54 &
9.43 & 5 & 18 &
49.10 \\

              & Planner &
\textbf{28.88} & \textbf{18.05} & \textbf{43.10} &
\textbf{6.46} & \textbf{4} & \textbf{10} &
\textbf{38.36} \\
\hline
\Esw~South & Straight-to-Goal &
\textbf{21.37} & 10.75 & 54.64 &
\textbf{6.00} & 3 & 14 &
\textbf{29.29} \\

              & Planner &
26.82 & \textbf{10.44} & \textbf{47.55} &
7.29 & 3 & \textbf{13} &
32.95 \\
\hline
\end{tabular}
\end{table*}

Results for the real-world deployments are reported in Table~\ref{tab:deployment-results}. These show that the Planner outperformed the baseline in $2$ out of $3$ deployments. For MOGli, as in simulation, the Planner leads to a reduction in the mean duration of approximately $30$ minutes on average; the minimum and maximum durations across all transects are also reduced. The Planner yields a reduction in the path length that is statistically significant at the 5\% level ($p=0.041$ and rank-biserial 
$r=0.376$ under the nonparametric Mann-Whitney U test with $20$ samples for STG and $21$ for the Planner). However, given the number of transects recorded (see Table~\ref{tab:deployment-stats}), statistical significance was not obtained for other deployments and metrics. These field results should be treated as corroborating evidence alongside the larger-scale simulation study rather than as standalone proof of generality.
The Planner also performed better than STG for \Esw{} North in all metrics. For \Esw{} South, the mean values were slightly worse, while the minimum / maximum ranges still matched or improved over STG. %

\begin{figure}[t]
	\begin{center}
		\includegraphics[width=\columnwidth]{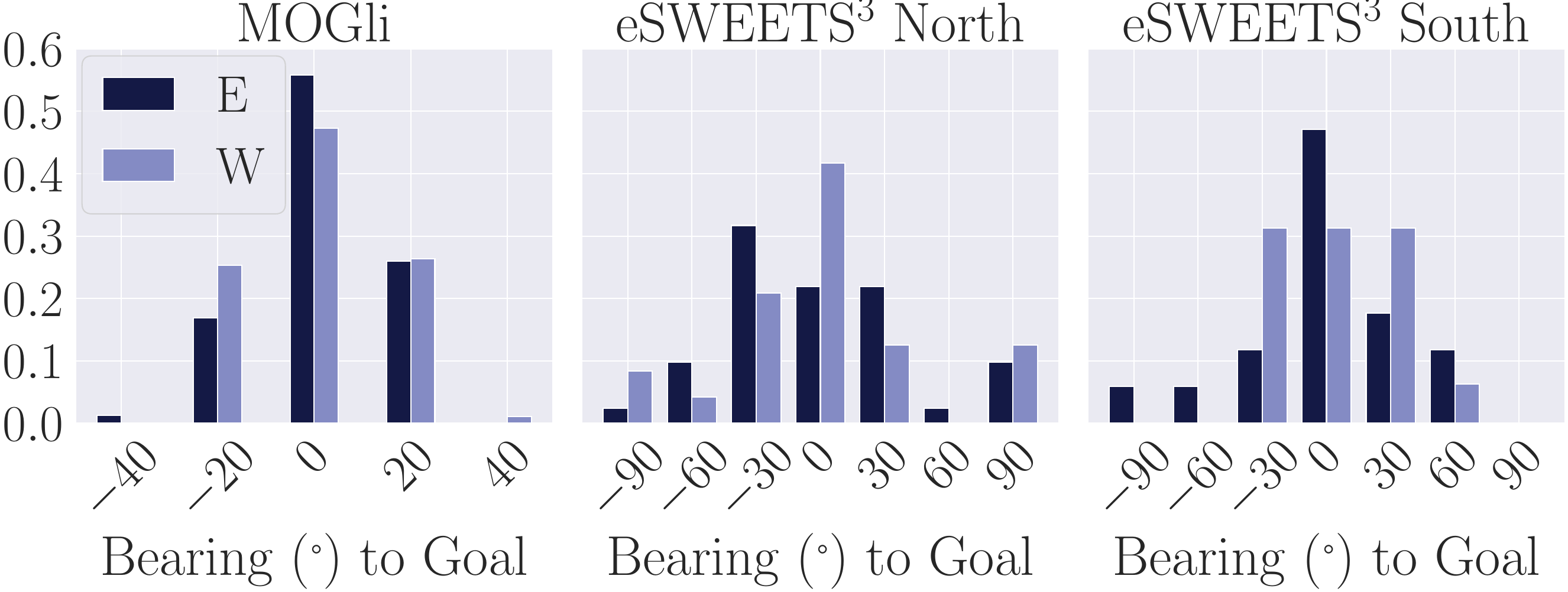}
		\caption{Proportions of actions chosen by the Planner during the three field deployments, split between the East and West directions of travel. The stronger currents of the \Esw~region resulted in a wider distribution of action frequencies.}
		\label{fig:action_frequencies}
	\end{center}
\end{figure}

In Figure~\ref{fig:action_frequencies}, we analyse the frequency of the actions taken by the Planner in the three deployments. While the STG action ($\alpha=0$) is the most selected overall, the Planner often chooses to steer relative to the goal. In the \Esw{} region, characterised by stronger tidal currents, the distribution of chosen actions is wider and flatter, and even the most extreme actions are still chosen $15\%$ of the time in \Esw{} North. 
The strong tidal currents in the \Esw{} region also resulted in surfacings outside of the safety polygon. 27 such surfacings (20.3\%) occurred for \Esw{} North and 4 (2.9\%) for \Esw{} South.

\subsection{Alignment of Simulation and Real-World Results}

For MOGli, the duration and number of dives are mostly aligned, while the path lengths are noticeably shorter in simulation. This is consistent with the distance underestimation observed in the simulator evaluation. 
For the \Esw{} deployments, values for all metrics were consistently higher in simulation compared to the field deployments. Given the values reported in Section~\ref{sub:simulator_evaluation}, simulator predictions of dive distance and durations are well-aligned with field observations. Consequently, the difference in transect metrics is attributable to the goal achievement conditions in the deployed autonomous system being broader (3 conditions, as described in Sections~\ref{sub:automated_goal_management} and~\ref{sub:deployments}) than those in simulation (1 condition). Requiring surfacing within a tight radius in simulation will result in overshooting the goal and hence longer transects until the goal is eventually achieved. Concretely, in the \Esw{} field trials, the goal was deemed achieved by reaching it underwater in $46\%$ of cases, having traversed 90\% of the transect length prior to safety polygon re-entry in $22\%$ of cases, and by surfacing within the goal radius $\rho$ in only $32\%$ of transects.

\subsection{Summary of Results}

Overall, our results demonstrate that the proposed online planning techniques outperform the standard Straight-to-Goal baseline. Improvements of up to 9.88\% in dive duration and 16.51\% in path lengths were obtained in simulations calibrated on real-world data. Our system also obtained a statistically significant path length reduction of 9.55\% in a field deployment. Compared to prior studies that are limited to simulation or short-term field trials, these results provide evidence of performance gains in real-world operational campaigns. A key achievement of our work is the demonstration of sustained autonomous operation at scale. The GALE system was deployed in two operational campaigns totalling over 3 months and 1000 km of autonomous glider operation, successfully addressing the challenges of environmental and actuation uncertainty, intermittent communication, and long-horizon navigation planning. 

\section{DISCUSSION}

\subsection{Lessons Learned and Engineering Considerations}

The feedback of pilots involved in the two campaigns highlighted the value of the system in automating standard operation. The system reduced manual effort and allowed pilots to focus on monitoring and non-routine interventions for other campaigns. In the \Esw{} campaign, the safety polygon feature allowed for automated intervention when pilot supervision may not have been available otherwise. Additionally, the straighter transects attained by our planning approach lead to more consistent sampling along predefined paths, which supports robust environmental data analysis and aligns with common scientific objectives.

The early stages of the project built gradually towards achieving the full control loop. For testing the systems integration, a ``shadow'' glider was initially configured. This glider duplicated the status updates received from the real glider. The control instructions generated by GALE were placed in a special folder where they could be sanity-checked by pilots but not acted upon. Subsequently, pilots began transferring the validated control instructions manually to the real glider. Once confidence in the system was achieved, fully autonomous piloting by GALE was enabled during working hours, with manual piloting outside of this window. By the start of the MOGli deployment, pilots were comfortable with simply monitoring the operation of the fully autonomous system regularly.

Regarding the generality of our system, given the modular architecture, \sysshort is extensible to other glider designs such as the Seaglider~\cite{eriksen2001seaglider}. From a software engineering standpoint, this would require replacing the SFMC and SFMC Adapter with vehicle-specific piloting systems and adapters. From a methodological point of view, while Seagliders have slightly different low-level actuators, they also require the specification of waypoint and control parameters in an online piloting loop. Thus, our methodology is applicable to other glider designs with minimal modifications.

\subsection{Challenges and Future Work}

We aim to generalise our method to also consider the optimisation of energy use, which is a feasible extension of the proposed formulation. As the duration and energy use objectives conflict, a multi-objective approach~\cite{roijers_surveymultiobjective_2013} is required. This would allow the selection of more energy-intensive controls during periods with strong currents in order to make sufficient progress towards the scientific objectives. Extending the method to incorporate constraints and guarantees, such as staying within a specified survey area or avoiding certain areas (e.g., shipping lanes) would broaden the applicability of the method. The problem may also be framed as a POMDP~\cite{sunbergOnline2018} in order to maintain beliefs about the nature of forecast errors, which would require in-situ current measurements. 

Replacing or augmenting the simulator with a learned model, which has proven successful with a variety of planning and reinforcement learning approaches~\cite{hafner2019learning,schrittwieser2020mastering}, may speed up execution and thus enable planning over a longer horizon. As shown by our simulator evaluation, further improvements in accuracy are possible. 
However, fitting simulator behaviours using different vehicles and deployment environments is a significant challenge due to distribution shifts. Further developments to the planner can include the aggregation of continuous actions for root-parallel MCTS~\cite{xiao2025gaussian}, improving the exploration strategy~\cite{painter2023monte}, and contextually deciding which actions to prioritise.

Evaluating glider navigation planning algorithms in the field is inherently challenging given the long traversals and, consequently, the low number of transects. If resources permit, running algorithms on gliders deployed side-by-side in a paired experimental design would reduce variability and temporal dependencies compared to the alternative-control design we adopted. Regardless, replaying on-field conditions in suitably calibrated simulators, as performed in our evaluation, is ultimately necessary for statistically robust findings.

Failures in transferring instructions to the glider also pose a unique challenge, as they can cause the glider to turn around towards a stale waypoint. Such failures disproportionately impact navigation plans with waypoints that steer away from the goal to compensate for currents. This phenomenon highlights the trade-off between the gains of online planning for steering given up-to-date current forecasts versus the negative impacts of using a stale online plan due to failed instruction transfers. This can be addressed by explicitly reasoning about the time available for planning~\cite{budd2024stop} and limiting it in order to improve the probability of instruction transfers succeeding.

In terms of applications, we envisage leveraging the present work as a building block. The system can operate in regions with high eddy activity using high-resolution ocean forecast models, a task that is challenging for current piloting methods. Identifying ocean fronts from horizontal temperature gradients~\cite{zhangAutonomousTrackingOceanic2019} or tracking algal blooms~\cite{fonsecaAlgalBloomFront2021} can be achieved by creating a higher-level component that reasons about where the highest-value observation(s) can be gathered and tasking the glider accordingly. Decision-making could also be implemented to minimise biases identified in existing sampling strategies (e.g.,~\cite{patmore2024evaluating}). For going beyond the single-glider scenario, coordination strategies can be considered in a multi-robot setting. High-level methods can be constructed that assign goals or regions to individual gliders, with our planner being used to compute uncertainty-aware navigation plans for each vehicle.

\section{CONCLUSION}

In this paper, we have considered the problem of navigation planning for underwater gliders, a type of robot that has become invaluable for ocean sampling. We have taken important steps to address the current methodological and systems gaps towards achieving long-term autonomous glider operation. Our work proposed a glider simulator and a procedure for adjusting its parameters using historical dives performed in the real world. We have also contributed a sample-based online planning technique for deciding how to control the glider with the objective of minimising duration to reach the goal. Lastly, in collaboration with pilots and domain experts, we have designed the \syslong, which is suitable for long-term and multi-platform missions.

We have thoroughly validated our approach in simulation and through operational campaigns in the North Sea, in which our system was used to control gliders collecting data towards refining weather models and understanding the impacts of offshore wind farms. Our work improves the sampling of scientific data and paves the way towards the autonomous management of fleets of gliders at scale, reducing costs and allowing human pilots to focus on monitoring and non-routine interventions. Our contributions ultimately support advancements in marine science and policy.

\section*{ACKNOWLEDGMENTS}
This work was supported by the UK Natural Environment Research Council (NERC) under the UKRI TWINE programme (grant number \texttt{NE/Z503381/1}) and the Doctoral Training Partnership in Environmental Research (\texttt{NE/S007474/1}). It was also supported by the UK Engineering and Physical Sciences Research Council (EPSRC) From Sensing to Collaboration Programme Grant (\texttt{EP/V000748/1}) and the Innovate UK AutoInspect Grant (\texttt{1004416}). The authors would like to gratefully acknowledge the contribution of the UK Met Office and the University of East Anglia, who gave permission for the \sysshort{} system to be trialled in their glider campaigns. The \Esw{} glider campaign was funded by the UK NERC (grant number \texttt{NE/X004872/1}). Lastly, we acknowledge the use of the University of Oxford Advanced Research Computing (ARC) facility in carrying out this work (see \url{http://dx.doi.org/10.5281/zenodo.22558}). 

\section*{ADDITIONAL INFORMATION}
A patent application has been submitted on the work covered in this paper, which was first filed on 19 December 2024 under UK application number \texttt{GB2418678.5}.

\bibliographystyle{IEEEtran}
\bibliography{references.bib}

\vfill\pagebreak

\end{document}